# HERMES: a multi-agent framework for structured knowledge extraction from ultra-long documents in geoscience

Ziqi Song†[1], Zongyuan Xiang†[1], James G. Ogg[2,6], Bruce S. Lieberman[3], Gabi Ogg[5], Natalia López Carranza[3], Wen Du[4], Yufei Ye[1], Shuan Li[1], Zhong Peng[1], Shaoqi Yu[7], Juye Wei[1], Ying Zhou[1], Jieping Ye[1], Jiang Yang[1*]

[1]Research Center for Computational Earth and Space Science, Zhejiang Laboratory, Hangzhou, China
[2]School of Electrical & Computer Engineering, Purdue University, Indiana, USA
[3]Biodiversity Institute, University of Kansas, USA
[4] Department of Earth and Environmental Sciences, University of Illinois Chicago, Chicago, IL, USA
[5] Geologic TimeScale Foundation, Indiana, USA
[6] Zhejiang Deep-time Digital Earth International Research Center, Hangzhou, China
[7] Ant Group, China

† These authors contributed equally to this work.

*Corresponding author: Jiang Yang (cscyangj@zhejianglab.org)

## Abstract

Authoritative scientific knowledge in geoscience remains largely trapped in legacy monographs and historical literature, where unstructured text and complex layouts hinder computational access. We introduce HERMES, a scalable multi-agent framework that extracts structured data from ultra-long scientific documents. Using a coordinating large language model, HERMES integrates domain constraints, validation rules and evidence tracing within a unified document-level extraction process that incorporates parsed text, tables, figures and captions. Applied to the 55-volume *Treatise on Invertebrate Paleontology*, the system produced a structured database of 32,277 fossil taxonomic entities and 451,878 attributes, released online at https://treatise.geolex.org. Extraction performance remained stable across fossil groups (average F1 scores of approximately 0.90 for entities and 0.91 for attributes), improving per-volume efficiency approximately sixfold relative to the tested fully manual baseline. Evaluation in palaeomagnetism and geochemistry, conducted without additional model training, demonstrated transfer across distinct geoscience domains. This work provides a practical pathway to transform historical scientific literature into FAIR-oriented structured data, offering a sustainable infrastructure for data-intensive disciplines and large-scale knowledge integration.

## Introduction

The rapid growth of scientific literature has made it increasingly difficult to access, structure, and computationally reuse published scientific knowledge. This challenge is central to the broader transition toward data-intensive scientific discovery, in which scientific progress increasingly depends not only on new observations but also on the ability to organize, integrate, and analyze existing data at scale [1–3,59]. For disciplines that rely heavily on historical records and published observations, the systematic curation and repurposing of data from existing literature is especially important, as primary data acquisition may be constrained by high costs, limited accessibility, or the historical and non-repeatable nature of many natural processes 2,4,5. Consequently, automated systems capable of extracting, validating, and structuring knowledge from legacy literature are becoming essential infrastructure for transforming static scientific archives into computable, reusable, and evidence-traceable knowledge resources, thereby supporting a broader shift toward data-intensive scientific discovery 3,5.

Concurrently, large language models (LLMs) and large multimodal models (LMMs) have achieved substantial progress in reasoning, coding, scientific question answering, and multimodal understanding. Representative foundation models, including GPT-4, Gemini 1.5, DeepSeek-R1, and Qwen2.5, have demonstrated increasingly strong capabilities in long-form reasoning, multimodal information processing, structured data analysis, and tool-oriented task execution [6–9]. Building on these models, agentic AI systems have begun to move beyond single-step question answering toward coordinated scientific tasks in which LLMs act as planners, controllers, and interfaces to external tools, databases, and computational environments [10–13].

Recent work further indicates that agentic AI is becoming a practical paradigm for scientific workflows. Multi-agent systems have been used to automate components of scientific discovery, including hypothesis generation, experimental planning, data analysis, and follow-up interpretation [14]. Related systems such as Co-Scientist and The AI Scientist show that LLM-based agents can assist with structured scientific thinking, literature synthesis, experimental design, code generation, result analysis, and even end-to-end automation of AI research [15,16]. Other agentic systems have targeted expert-level empirical software development and traceable clinical reasoning for rare disease diagnosis, demonstrating that tool-using agents can operate in specialized, evidence-intensive domains [17,18]. These developments establish the broader relevance of multi-agent architectures for scientific work, but they do not directly address the large-scale, evidence-traceable extraction of structured knowledge from legacy geoscience literature.

The technical foundation for such systems has been shaped by advances in reasoning–acting frameworks, tool use, and multi-agent coordination. ReAct introduced an interleaved reasoning-and-acting paradigm, enabling LLMs to generate reasoning traces while interacting with external environments [10]. Toolformer showed that language models can learn to decide when and how to invoke external tools through self-supervised API-use signals [11]. Multi-agent frameworks such as AutoGen and CAMEL further demonstrated that complex tasks can be decomposed into interactions among specialized agents with distinct roles, memory, and communication patterns [12,13]. More recently, protocol-oriented efforts such as the Model Context Protocol and Agent2Agent protocol have begun to define standardized interfaces for connecting agents with tools, data sources, and other agents [19,20]. These developments provide a conceptual and technical basis for designing modular scientific extraction systems in which different sub-agents specialize in parsing, retrieval, validation, and provenance tracking.

Scientific information extraction has evolved from task-specific entity and relation extraction toward document-level and LLM-based structured extraction. Early scientific IE benchmarks and datasets such as SciERC and SciREX established entity, relation, coreference, and document-level extraction tasks for scientific papers, while large-scale corpora such as S2ORC enabled broader research on machine-readable scholarly text [21–23]. In parallel, domain-specific systems and models in chemistry and materials science, including ChemDataExtractor, MatSciBERT, and recent LLM-based structured extraction frameworks, have shown that scientific text can be converted into structured, machine-actionable records [24–27]. These studies demonstrate the feasibility of scientific literature extraction, but most focus on journal articles, short passages, or relatively standardized text, leaving ultra-long monographs, complex layouts, and fine-grained evidence tracing insufficiently addressed.

Geoscience provides a stringent testbed for these limitations. Key information such as sampling coordinates, stratigraphic ages, fossil taxonomic names, palaeomagnetic directions, geochemical compositions, and analytical metadata is often distributed across narrative descriptions, tables, figure captions, appendices, and supplementary files. Unlike many benchmark extraction tasks, geoscience literature frequently requires linking entities to attributes across heterogeneous document regions, normalizing domain-specific units and nomenclature, and preserving uncertainty markers that carry scientific meaning. As a result, geoscience literature extraction requires not only entity recognition and attribute extraction, but also layout-aware parsing, cross-table reasoning, domain-specific validation, and evidence-level provenance [2,3,5,28]. Few existing systems integrate these requirements into a single deployable framework for large-scale, auditable extraction from ultra-long geoscience literature.

A first technical challenge is document length. Although LLMs can perform well on short scientific passages, full articles, monographs, and book-length documents pose substantially harder problems. Retrieval-augmented generation addresses part of this challenge by retrieving task-relevant passages from an external corpus before generation [29]. Long-context models and fine-tuning methods, such as Gemini 1.5 and LongLoRA, have extended the context windows available to LLMs, enabling models to process much longer inputs than earlier systems [7,30]. However, long-context capability alone does not guarantee reliable information use. "Lost-in-the-middle" analyses show that models often underuse information located in the middle of long contexts, and LongBench demonstrates that long-context understanding remains difficult across question answering, summarization, retrieval, and code-related tasks [31,32]. These limitations are especially problematic for scientific monographs, where relevant information may be sparsely distributed across hundreds or thousands of pages.

A second challenge is verification. For scientific data extraction, correctness depends not only on whether an entity or value is extracted, but also on whether it can be traced back to precise evidence in the original document. RAG systems partially address this issue by conditioning generation on retrieved passages, but provenance, attribution, and citation faithfulness remain open problems [29,33]. Recent work on context attribution, such as ContextCite, seeks to identify which parts of the input context are responsible for model-generated statements, while human-in-the-loop information extraction studies show that expert review can improve both efficiency and trust in extracted data [33,34]. In geoscience data production, paragraph-level or document-level attribution is often insufficient: experts need to verify values against exact sentences, table cells, figure captions, or page regions. This motivates evidence-aware extraction processes that combine structured outputs with fine-grained provenance and human validation.

To address these challenges, we introduce HERMES (Hybrid Entity-Relation Multi-agent Extraction System). HERMES integrates natural language processing, computer vision, entity recognition, and human-in-the-loop validation to automatically extract and structure multimodal data from geoscience literature. The system follows the ReAct architecture [10], with a geoscience LLM serving as the orchestrator that coordinates and dispatches specialized sub-agents, each responsible for a distinct aspect of data extraction. Together, they perform document digitization, entity and attribute extraction, format standardization, and evidence tracing. A shared context connects these sub-agents, allowing parsed text, tables, figures, intermediate entities, extracted attributes, validation results, and evidence locations to be updated and inspected throughout the coordinated extraction process. While a simpler pipeline architecture might handle tasks such as text or figure processing, HERMES offers advantages in modularity, parallelization, and error handling. For example, we can independently develop a table recognition algorithm for the complex table structures common in geoscience literature, simultaneously extract and trace multiple entities or attributes in parallel, or replace the orchestrator with another LLM to adapt the system to other disciplines. Our architecture is consistent with recent advances in multi-agent coordination, where the orchestration role improves overall performance and supports scalability in multi-task processing systems.

HERMES strikes a balance between cost and performance. The geoscience LLM has 72 billion parameters and is deployable on a single server with 150 GB of GPU memory to run long-context inference at FP16 precision. In the palaeontology case study, HERMES achieves high entity and attribute extraction performance while substantially reducing the manual effort required for large-scale curation. This underscores the benefits of decomposing automated information extraction into smaller tasks, each executed by a coordinated, task-specific sub-agent, and highlights the importance of agent collaboration.

Using HERMES, we have digitized the *Treatise on Invertebrate Paleontology* (TIP)[37], and organized the results into an openly accessible database. The TIP is the most authoritative and foundational literature series in palaeontology on the taxonomy, morphology, classification, stratigraphy, and geography of tens of thousands of fossil and modern taxa, containing both images and text in more than 50 volumes. Usefully, the text in the TIP has been fairly consistent over the course of seven decades, albeit with some variation across volumes, making it an ideal test case to generate a database that combines several different types of information at different hierarchical levels. The database created contains 32,277 fossil genus-level entities and 451,878 attribute records, and was constructed in only 53 person-days. This achievement makes palaeontological knowledge, previously scattered across dozens of monographs far more FAIR, and now accessible and analyzable in a computable, searchable, and reusable form. The resulting database can also be continuously updated as additional TIP volumes are published and processed. Furthermore, we have validated the cross-disciplinary capability of HERMES on palaeomagnetic and geochemical literature. Our results offer promise for accelerating research in geoscience by streamlining literature based data collection and curation.

# Results

Large-scale, data-driven, and interdisciplinary scientific analysis requires methods to automatically extract structured data from the vast and dispersed body of published literature. Geoscience provides a stringent testbed for this problem because relevant information is often distributed across legacy monographs, scanned documents, narrative descriptions, figures, captions, and complex tables. To address this, we developed and used HERMES, a multi-agent system designed to transform scattered information from scientific publications into structured, high-fidelity, and evidence-traceable datasets.

The system is built upon a multi-agent architecture implementing the ReAct paradigm[10]. A central Orchestrator agent conducts high-level reasoning (Fig. 1): it interprets user queries, assesses task state, formulates plans, and coordinates a suite of specialized sub-agents. Each sub-agent—Parser, Entity Recognizer, Annotator, Validator, and Tracer—is optimized for a distinct phase of the extraction process. Together, these sub-agents separate document parsing, entity recognition, attribute extraction, rule-based validation, and evidence tracing into coordinated but inspectable modules.

This coordinated design establishes a transparent and adaptable framework for automated knowledge extraction. Furthermore, the framework supports human-in-the-loop validation, where expert-adjudicated data can be written back to the shared context to refine the extraction process and improve subsequent validation, a strategy that has been shown to increase efficiency and trust in information extraction systems[35].

## The Coordinated Multi-Agent Extraction Process of HERMES

The system process (Fig. 1) begins with a user uploading scientific literature in various formats (e.g., scans, digital manuscripts), which are standardized as PDFs, and providing a natural language query. This query defines the entities of interest and, optionally, their specific attributes for extraction.

The central Orchestrator agent first establishes a shared context to coordinate all subsequent steps. It directs the Parser agent to digitize the PDF, extracting raw text, figures (together with their captions), and tabular data into the shared context. Employing the user query and the Parser output, the Orchestrator then invokes the Entity Recognizer to identify all relevant entity mentions, recording each entity name and its sentence-level source.

For attribute extraction, the Orchestrator calls the Annotator. To ensure efficiency with long documents, the Annotator employs a retrieval-augmented generation (RAG) approach[29]: for each target attribute, the Annotator retrieves the most relevant text chunks and writes the extracted value and its source evidence to the shared context. All extracted data is then passed to the Validator agent for domain-rule-based consistency checking and formatting.

Finally, the Orchestrator activates a Tracer agent, which records the source evidence and its page-level bounding-box coordinates within the original document in the shared context. This design is aligned with recent efforts to improve attribution and verifiability in retrieval-augmented and context-grounded generation systems [33].

In the experiments reported here, GeoGPT-R1-Preview[36], a geoscience-oriented model based on Qwen2.5-72B, served as the Orchestrator and was used by the Annotator for attribute extraction, whereas DeepSeek-V3[58] was used by the Entity Recognizer for entity extraction.

The same model assignment was used across palaeontology, palaeomagnetism and geochemistry. Model weights remained unchanged; only the extraction schemas and domain-specific Validator rules were adapted.

The ReAct-based architecture, facilitated by the dynamic shared context, makes this multi-agent process inherently adaptable to diverse disciplinary requirements, as demonstrated in its application to palaeontology, palaeomagnetism and geochemistry detailed in the following sections.

## Palaeontology: data extraction from *Treatise on Invertebrate Paleontology*

The *Treatise on Invertebrate Paleontology* is a major, ongoing scholarly work that has been published and updated since 1953. It synthesizes research from hundreds of global experts across taxonomy, morphology, stratigraphy, and palaeobiogeography, comprising 55 published volumes to date [37]. Relative to the textual scale of each volume (averaging 160,437 words), descriptions of individual fossil entities are very brief (averaging 97 words). This characteristic of highly sparse and dispersed information renders the cost of manually identifying fossil entities and extracting their multi-dimensional attributes prohibitively high. Existing palaeontological databases demonstrate the scientific value of structured fossil data, but they also illustrate the dependence of palaeobiological data resources on extensive expert curation [38]. This has limited the application of data-driven research paradigms in palaeontology.

We employed HERMES to construct a comprehensive fossil database from the *Treatise on Invertebrate Paleontology*. The task was performed volume-by-volume, with inputs including the document and a user query (see Supplementary Information). As shown in Fig. 2, the Orchestrator first invokes the Parser to process the input volume. Subsequently, the Orchestrator filters all fossil-relevant passages from the parsed output: first, via explicit rules, such as identifying taxonomic name patterns, author–year citations, and uncertainty markers used in zoological nomenclature [39], and second, by applying its geoscience expertise to perform a secondary judgment on candidate passages.

The Orchestrator then calls the Entity Recognizer to extract fossil entities from these identified passages. To balance processing efficiency and extraction precision, the Orchestrator processes three fossil-relevant passages at a time. Next, the extracted fossil entities and their corresponding source passages are passed to the Annotator, with a strict constraint that attribute extraction is confined to the provided passages. This constraint is crucial to prevent the Annotator from hallucinating information beyond the immediate context. Following the user query, the Annotator extracts attributes including taxonomic information, morphological characteristics, geological age, geographic distribution, and photographic plate references, outputting the results in JSON format. The Validator applies deterministic rules (see Supplementary Information) derived from domain knowledge and community standards, including zoological nomenclature and chronostratigraphic conventions [39,40]. These rules standardize fossil-name structures, preserve uncertainty markers, unify caption and citation formats, and decompose or recombine geological-age fields.

Finally, the Orchestrator invokes the Tracer to identify the source evidence for each entity and attribute. In subsequent manual review, mapping these evidence locations back to the original PDF pages for highlighting provides intuitive support for result verification.
After manual verification, extraction results were released on the Treatise.geoLex database[41], which now contains 32,277 taxonomic entities and 451,878 attributes. A team of eight evaluators, including palaeontologists from Purdue University, conducted the manual verification. As new volumes of the Treatise are published, they will be processed using the same extraction framework and incorporated into the database through continuous updates.

The HERMES workflow, consisting of automated data extraction followed by manual review and revision, achieved a mean end-to-end turnaround time of 7 calendar days per volume. For comparison, a fully manual extraction experiment conducted on one volume by a team of four annotators required 45 calendar days, corresponding to an approximately sixfold reduction in per-volume turnaround time. Across the complete 55-volume database-construction process, the recorded cumulative labour input was 53 person-days. Beyond the measured reduction in turnaround time, domain experts involved in data verification noted that, without automated pre-extraction, they would not have considered undertaking comprehensive extraction of the complete Treatise series. HERMES therefore not only accelerated the processing of individual volumes but also made the construction of a series-wide structured database practically feasible.

To evaluate palaeontological extraction, the original, unedited HERMES output was archived before expert review. Domain experts subsequently inspected the source documents and used the system output as an initial annotation layer, correcting erroneous records, removing spurious records, and adding entities or attributes missed by the system. The resulting expert-adjudicated reference dataset was used for evaluation. Evaluation was exhaustive within all volumes represented in Table 1, and no record-level subsampling was performed. Entity- and attribute-level metrics were calculated by comparing the archived pre-review output with the final expert-adjudicated reference dataset.

Performance of HERMES in this task is shown in Table 1. Across the evaluated fossil groups, entity recognition achieved an overall recall of 0.89, precision of 0.90, and F1 score of 0.90. Attribute extraction achieved an overall recall of 0.87, precision of 0.95, and F1 score of 0.91, indicating that the system maintains strong performance even when extracting semantically richer and more heterogeneous attribute information. Performance was broadly stable across fossil groups, despite substantial differences in document length, and some modest differences in layout structure and taxonomic terminology. These results suggest that HERMES does not rely solely on local context or templated patterns, but can locate, associate, and reconstruct entity–attribute information within the broader context of each taxonomic group considered by Treatise volumes.

## Palaeomagnetism: cross-domain validation

Unlike palaeontological monographs, which are primarily taxonomic narratives, palaeomagnetic literature centers on quantitative measurements and experimental parameters, consistent with the structured data models used by community palaeomagnetic archives such as MagIC [42]. As shown in Fig. 3, a substantial proportion of key attributes are presented in tabular form, including large-scale datasets that span multiple pages or interlinked tables. This represents a class of scientific text dominated by numerical attributes and table-driven structures.

For this task, we adapted the validation rules within the Validator agent for palaeomagnetic data. Each extraction run takes as input a single document along with a user-defined query (see Supplementary Information), and produces hierarchical entity–attribute outputs, with each site or sample entity linked to its extracted measurement, age, polarity, and locality attributes.

HERMES was applied to a corpus of 567 palaeomagnetic articles. Because construction of document-level expert-adjudicated reference datasets required complete manual review of the original publications, quantitative evaluation was conducted on 13 articles selected by domain experts for full annotation. Article selection was completed before metric calculation and was not based on extraction performance. For each selected article, all entities and attributes were evaluated by comparing the original HERMES output with the final expert-adjudicated reference dataset. The remaining articles were processed as part of the data-production workflow but were not included in the reported accuracy estimates because complete expert-adjudicated reference datasets were unavailable.

As shown in Table 2, entity extraction achieved high performance, with a precision of 0.93 and a recall of 0.96. Attribute extraction, which requires linking numerical values to the correct sites, samples, and measurement fields, achieved a precision of 0.81 and a recall of 0.77. These results show that HERMES can handle palaeomagnetic literature in which structured information is frequently embedded in numerical tables and multi-field measurement records.

## Geochemistry: cross-domain validation

As shown in Fig. 4, geochemical research places greater emphasis on sample-level records and high-dimensional attribute organization. Each sample is typically associated with geographic location, tectonic setting, rock name, age, and multiple major, trace, and isotopic indicators, forming a typical one-entity-multiple-attribute data structure. This high-dimensional structured knowledge extraction task requires the system not only to recognize sample entities but also to extract and bind many numerical attributes along with their metadata reliably.

We used the GEOROC data system[43,44] to evaluate the performance of HERMES. GEOROC is a globally recognized, peer-reviewed geochemical database that has long compiled whole-rock and mineral compositional data for igneous and metamorphic rocks and related minerals from tens of thousands of publications through manual curation. The test set was derived from the dataset of Ball[45], which reports the global distribution and composition of Neogene to Quaternary intraplate volcanic rocks.

For this test, we adjusted the validation rules related to geochemical field formats within the Validator, while leaving all other system configurations unchanged. Each extraction run took a single document and a user-defined query as input (see Supplementary Information) and produced structured, entity-centered outputs in which each sample entity was associated with its extracted geochemical attributes.

As shown in Table 3, entity extraction in geochemical literature achieved an overall recall of 0.89, precision of 0.74 and F1 score of 0.81. Attribute extraction achieved a recall of 0.35, precision of 0.93 and F1 score of 0.51. The source geochemical tables contained at least 67 variables, from which domain experts selected 10 target attributes for the present evaluation. HERMES therefore did not attempt to extract every variable in the source dataset; instead, it was required to locate and associate the selected attributes within wide tables containing numerous additional columns, multi-level headers, units and footnotes. The reduction in attribute recall indicates that locating and binding selected values within such high-dimensional table structures remains challenging, particularly when target attributes are distributed across multiple tables or when their units and analytical context are specified only in headers or footnotes.

In this scenario, missing attributes are typically not caused solely by entity recognition errors. For example, when certain element or isotopic indicators appear only in supplementary or cross-page tables, or when their unit information is implicitly given only in the table header, the system may successfully identify the sample entity but fail to extract all relevant attributes, thereby lowering overall recall. Nevertheless, the high attribute precision indicates that, under high-dimensional data structures, HERMES tends to prioritize correct alignment between numerical values and entities rather than generating uncertain candidate values.

# Discussion

Scientific literature contains extensive observational and descriptive knowledge, but much of this information remains difficult to convert into needed FAIR data. This limitation is especially pronounced in data production settings where relevant evidence is dispersed across narrative text, tables, figures and captions, and where extracted values must remain traceable to their original sources. Here we introduced HERMES, a multi-agent framework that organizes document parsing, entity recognition, attribute extraction, validation and evidence tracing into a scalable extraction process. Validation across palaeontology, palaeomagnetism and geochemistry shows that HERMES can maintain robust performance across distinct document structures and data types. These results establish HERMES as a reusable framework for transforming scientific literature into structured, computable and evidence-traceable data resources. When clear entity and attribute structures can be defined and basic domain constraints are available, HERMES can be adapted to other scientific literature domains to support large-scale, sustainable data construction and updating.

The two-stage entity-attribute extraction strategy adopted in HERMES enhances system controllability and auditability but also introduces potential error propagation. The entity index constructed during entity recognition serves as the foundation for attribute extraction. If an entity was missed, a boundary error was made, or homonymous entities were merged improperly, the attribute extraction results can be omitted or incorrectly associated, leading to cascading errors in the output structure. As shown in Fig. 5(a), a fossil description paragraph for "*Platycoryphe*" provides only partial information, lacking figure captions and characteristic descriptions. Incomplete records can occur regularly in monographs. During paragraph filtering, HERMES identifies fossil-relevant passages using cues such as taxonomic names, plate numbers, and reference identifiers. Passages lacking these cues are excluded to reduce false positives, preventing unrelated text from being recognized as independent fossil entities and avoiding redundant extraction of associated attributes. However, this conservative filtering strategy can also reduce recall when genuinely relevant fossil descriptions are incomplete or expressed without these cues.

When processing extremely long documents, HERMES relies on strategies such as rule-based filtering, text chunking, and RAG for attribute extraction to reduce computational complexity[29]. However, this dimensionality reduction inevitably introduces a trade-off between efficiency and recall. This limitation is consistent with broader findings that long-context models may still fail to use information reliably when relevant evidence is sparsely distributed or located in less salient parts of the context [31,32]. In some cases, information is scattered across non-contiguous pages, figures, or captions. When optical character recognition (OCR) results are segmented during layout analysis, originally continuous text can be divided into multiple incomplete fragments, reflecting a known challenge in document layout analysis and reading-order reconstruction [46]. As shown in Fig. 5(b), a fossil description spanning two pages is parsed as two separate text segments, with the first containing the entity name and the second containing attribute information. In such cases, if filtering rules operate on individual lines or segments, the latter part of the split description can be excluded from the subsequent extraction process, resulting in reduced recall in the affected passages. This issue is particularly pronounced in extremely long documents or complex layouts.

Complex table structures remain a major challenge in automated extraction, as also shown by recent table extraction and table reconstruction benchmarks that emphasize the difficulty of recovering complex headers, merged cells, and functional table structures from scientific documents [47,48]. Although HERMES can recognize and parse most tabular data in scientific literature, fully understanding table semantics remains difficult in cases involving cross-page tables, multi-table relationships, complex row/column mergers, and footnote information. For example, in Fig. 5(c), sample-level metadata and compositional measurements are stored in separate tables. One table records sample IDs together with formation, borehole, depth, thickness, total organic carbon (TOC), and sorption-related properties, whereas another table reports mineral compositions for the same sample IDs. Establishing complete sample profiles therefore requires cross-table reasoning based on shared sample codes. Consequently, tables with complex layouts or non-standard header structures increase the risk of errors in binding entities to their attributes.

The overall performance of HERMES is influenced by the underlying OCR and layout analysis algorithms. Scanned images, complex layouts, and special symbols (e.g., Latin binomials or stratigraphic age markers) in scientific literature can introduce recognition noise, and OCR errors have been shown to affect downstream retrieval and information extraction tasks[49]. This problem is more pronounced in older publications, which are often scanned from photocopies and converted to PDFs, resulting in lower image resolution and blurred or broken character edges that cause misrecognition. For example, in Fig. 5(d), the fossil name "*Beraunia*" was recognized as "*Bcraunia*" due to blurred edges of the letter 'e'. Such issues introduced during prior scanning of older text documents can prevent entity names from matching standard names in existing databases. In contrast, newer born-digital documents typically offer more stable recognition quality. Although the system corrects some OCR errors through rule constraints and format normalization, underlying recognition mistakes can still propagate into downstream extraction results.

HERMES adopts the "automatic extraction + human verification" model to ensure that data meets publication-grade quality standards. However, the human verification step still relies on a limited pool of domain experts. In large-scale data production scenarios, the throughput of this step represents a practical bottleneck. This reflects a broader trade-off in human-in-the-loop information extraction: expert review improves trust and data quality, but its cost and availability constrain scalability[34,35]. Further exploration of more efficient quality control mechanisms is needed to enable HERMES to operate in highly automated, large-scale data production environments.

# Methods

## Orchestrator

In the experiments reported here, we used GeoGPT-R1-Preview as the Orchestrator agent, providing the system with core capabilities in semantic understanding, reasoning, and task orchestration. GeoGPT-R1-Preview is a large language model based on the Qwen2.5-72B foundation model[9,36], featuring strong reasoning capabilities in answering geoscience questions. In HERMES, the Orchestrator interprets the user query, determines task state, decomposes the extraction process into executable steps, and dispatches specialized sub-agents according to the information required at each stage.

Unless otherwise stated, all LLM inference settings used in HERMES are summarized in the Supplementary Information.

Although GeoGPT-R1-Preview served as the controller in this implementation, HERMES supports task-specific LLM backends for individual sub-agents. In the reported experiments, the Entity Recognizer used DeepSeek-V3[58] for entity extraction, whereas the Annotator used GeoGPT-R1-Preview for attribute extraction. This model assignment reflects the experimental configuration rather than a framework constraint, and the model backend of each sub-agent can be replaced independently.

## The Shared Context

The Orchestrator maintains a shared context for each document to manage the intermediate results generated during collaboration among sub-agents. The shared context is a document-level data structure that consolidates multimodal content from the source literature. All sub-agents exchange data exclusively through this shared context. This centralized information architecture enables HERMES to preserve semantic consistency across different processing stages and supports result auditing and process reproducibility.

The shared context is summarized schematically in Fig. 1. It contains the following information: digitized content from the PDF document (text, figures and captions, tabular data, and layout information); paragraph and chunk structures; entity indices; entity-level attribute records; and field-level evidence provenance with validation flags. This structure was directly reused for data extraction tasks in palaeomagnetism and geochemistry and can be adapted for other disciplines with minimal modification.

## Parser

The Parser agent digitizes raw content from PDF documents, including text, figures, and tables. Internally, it employs the Mathpix OCR engine[50] and an in-house layout analysis module informed by document layout analysis methods[46] for text recognition, generating outputs that preserve page numbers and spatial coordinates. Based on these outputs, the Parser sorts the recognized elements to restore paragraph boundaries and reading order, with particular attention to multi-column layouts and mixed text-figure compositions, ensuring that the extracted text adheres to the original semantic sequence.

For figures, the Parser not only extracts the image regions themselves but also captures their accompanying captions. Additionally, it searches the full text for sentences referencing each figure, performing figure-text matching based on figure numbers, spatial proximity, and semantic cues.

For tabular data, the Parser incorporates a table recognition algorithm developed in-house[48], specifically optimized for the complex table structures commonly found in geoscientific literature. This design is motivated by the known difficulty of recovering complete table structure, headers, merged cells, and functional relationships from unstructured scientific documents[47]. It generates recognition results in both LaTeX and HTML formats.

All parsed outputs are written to the shared context, serving as the exclusive input source for the Entity Recognizer and Annotator agents.

## Entity Recognizer

The Entity Recognizer extracts target entities specified by the user query from the source documents and builds an entity index. Each target entity in the user query is described by a name, a definition, and an example.

The Entity Recognizer employs the Llama-Index framework[51] for building and querying the retrieval index. First, it segments the text from the shared context into text chunks and encodes each chunk using bge-base-en-v1.5[52] to construct a full-text retrieval space. Text chunking operates at the sentence level, with each chunk containing at most 1024 tokens and an overlap of 128 tokens between adjacent chunks to reduce the risk of truncating entity names or key descriptions at chunk boundaries.

Next, the Entity Recognizer retrieves text chunks most likely to contain the target entities from the full-text retrieval space. Retrieval proceeds in two steps. In the first step, the Entity Recognizer combines BM25 keyword matching[53] with vector semantic similarity to recall up to 40 candidate chunks. The default scoring weight for both BM25 and vector similarity is 50%. Adjusting the scoring weight can adapt HERMES to different types of literature. For example, the scoring weight of BM25 keyword matching can be increased in terminology-heavy scenarios, whereas the vector similarity weight can be increased in contexts that rely on semantic understanding. In the second step, the Entity Recognizer uses bce-reranker-base_v1[54] to re-rank the relevance between the target entity description and each candidate chunk, selecting the ten most relevant chunks.

Finally, the Entity Recognizer calls DeepSeek-V3 to extract entity names that appear explicitly in these chunks, along with a brief description for each entity. This approach effectively reduces redundant output from the large language model.

When the user requests extraction of only the target entities without their associated attributes, the Entity Recognizer outputs each entity name together with a concise description (no more than 150 words) as supplementary information for that entity. When the user requests simultaneous extraction of both target entities and their associated attributes, the Entity Recognizer excludes content from references and footnotes, returning only entity names mentioned in the main body of the text.

Extracted results are normalized and mapped to unique entity identifiers before output. The Entity Recognizer resolves ambiguous or non-unique entity mentions by incorporating contextual information into the normalized entity name. For example, in some petroleum geology publications, pits or sampling points are labeled only with numerals such as “1”, “2”, and “3”, while multiple regions in the same document may each contain identically numbered entities. To avoid conflating these mentions, the system augments the original name with the relevant regional or geographic context, normalizing “1” to forms such as “Region A–1” or “Basin X–1”. This process ensures that each entity is uniquely represented within the document.

## Annotator

According to the entity index, the Annotator extracts the values of user-specified attributes for each entity.

Each attribute in the user query is described by a name, a definition, and an example. This description, together with the target entity's name and description obtained from the Entity Recognizer, forms the query for attribute extraction. The Annotator retrieves the top-10 text chunks from the full-text retrieval space that are most similar to this attribute extraction query. It then instructs GeoGPT-R1-Preview[36] to extract attribute values that appear explicitly in these text chunks. If a given attribute is absent from the text, the model must return a unified missing indicator rather than performing inferential completion. Internally, the Annotator executes extraction tasks in parallel using five threads.

## Validator

The Validator ensures that extraction results conform to domain-specific scientific expression standards. It normalizes the format of extracted entities and attributes through rule-based constraints. In addition, the Validator performs logical consistency checks, such as the sequential order of temporal ranges and the hierarchical relationship between attribute values and entity types. Fields that fail validation are flagged and passed to the Tracer agent and the expert verification stage.

The constraints applied by the Validator include naming structures, regular expression patterns, and domain-specific term lists. The objects of these constraints include naming formats, numerical identifiers, units, and temporal ranges.

Taking the palaeontology data extraction task as an example, the Validator normalizes entities and attributes based on naming conventions, derived from zoological nomenclature standards [39]. Following fossil nomenclature rules, when an entity name consists of multiple words, only the first word is retained. For instance, when the text contains the name *Fusulina* sp., the Validator normalizes it to *Fusulina*. For forms that carry clear taxonomic meaning, the original expression is preserved. For example, when a name appears as an abbreviated genus followed by a dot and a parenthesized qualifier such as a subgenus (e.g., *C.* (*Costaclymenia*)), or when the first word itself is a generic abbreviation (e.g., *C.*), the Validator retains the abbreviated form to avoid misinterpretation of its taxonomic meaning. Regarding uncertainty markers common in taxonomic texts (fossil names prefixed with "?"), the Validator treats the marker as a valid part of the name and replaces the entity name in the entity list with the uncertainty-marked form. For example, when a fossil name begins with "?" (e.g., ?*Lingula*), the large language model may sometimes return only *Lingula*; the Validator ensures the marker is preserved and records the entity as ?*Lingula*.

Based on the field description formats prescribed in the monograph, for descriptive fields such as geological age, the Validator uses regular expression pattern matching to decompose the age information according to chronostratigraphic terminology[40]. Text outside parentheses is treated as the main age range, while text inside parentheses is extracted as a finer-grained stage. For example, when the text contains "Devonian (Frasnian)", the system parses "Devonian" as the main geological period and extracts "Frasnian" as the corresponding stage.

For caption information (repository details and citations) accompanying figures, the Validator performs structured parsing on the figure caption text for each entity to obtain a consistent structured attribute representation. Figure captions typically contain one of two types of information: a repository institution code or a citation expressed as author–year. For repository codes, the Validator maps them to full institution names using a pre-built lookup table of repository abbreviations (e.g., "NHMUK" is resolved to "The Natural History Museum,

London, UK"). If no repository code appears in the caption, the Validator directly extracts the citation information consisting of the author and a four-digit year (e.g., "Smith, 1987").

## Tracer

The Tracer agent matches each extracted entity name or attribute value to its corresponding source evidence in the document.

The full text is stored in the shared context as units of sentences or tables (referred to as "chunks" in this section). The Tracer adopts a coarse-to-fine strategy and returns the most relevant source evidence for each extracted item.

In the coarse stage, the Tracer identifies text chunks that may contain the target entity name or attribute value. On one hand, all text chunks that explicitly contain the target entity name or attribute value are selected as candidates. On the other hand, when an entity or attribute value no longer appears explicitly in the original text due to normalization, the Tracer performs selection based on semantic similarity.

Specifically, the Tracer uses the following template to wrap the entity name:
[entity name]: In the document, [User_query.entity.name] includes [entity name]

For attribute values, the Tracer uses the following template:
[attribute value]: In the document, [User_query.attribute.name] of [entity name] is [attribute value]

Taking the processing of one geochemical article[55] as an example, the user query is:
"
Entity:
    Name: sample
    Definition: the sample no. or experiment number
    Example: LC7, 7, LC-7
Attribute:
    Name: mineral composition
    Definition: Refers to the mineral elements and the content ratio of the elements in the mineral or test sample
    Example: Carbonate 18.7%, Illite 1.3%
"

An extraction result is: entity name = "bmf-1", attribute value = "quartz 18.00%, k-feldspar 7.10%, muscovite 1.30%, carbonate 16.70%, kaolinite 47.29%, illite 1.30%, chlorite 16.70%".

They are wrapped as follows:
bmf-1: In the document, the sample entity includes bmf-1.
quartz 18.00%, k-feldspar 7.10%, muscovite 1.30%, carbonate 16.70%, kaolinite 47.29%, illite 1.30%, chlorite 16.70%： In the document, mineral composition of bmf-1 is quartz 18.00%, k-feldspar 7.10%, muscovite 1.30%, carbonate 16.70%, kaolinite 47.29%, illite 1.30%, chlorite 16.70%.

Then, the Tracer uses the BGE-M3[56] model to generate embeddings of the two templates and each text chunk. Cosine similarity is calculated between each template embedding and each chunk embedding. The top 256 chunks with the highest similarity are selected as candidates for the next phase.

In the second phase, the Tracer uses the BGE-reranker-large[57] to rank all candidate chunks and returns the top-1 result. The text and coordinates of this text chunk are then written to the shared context.

## Human-in-the-loop Validation

HERMES provides native support for human-in-the-loop validation[34,35]. Users can review extraction results through a dedicated interactive workbench within the system. As illustrated in Fig. 6, the workbench displays the original PDF document on one side and the extracted entity-attribute results in a structured format on the other. When a user selects a specific entity or attribute, its source evidence is automatically highlighted in the original document. Users can revise entity names or attribute values via double-click editing. Benefiting from the two-stage entity-attribute extraction strategy, any correction made to an entity triggers an automatic re-extraction of its associated attributes. In the palaeontological monograph extraction task, palaeontology experts from Purdue University employed this workbench to validate approximately 2,000–3,000 entity records per month on average.

By positioning human validation as an architecturally decoupled stage that nevertheless forms a closed feedback loop with the automated extraction data layer, HERMES maintains overall scalability while delivering publication-grade data quality assurance.

## Generative AI assistance

During the preparation of this manuscript, the authors used ChatGPT (OpenAI) solely to improve language clarity, grammar, and readability. All scientific content, methodological design, analyses, interpretations, and conclusions were developed, verified, and approved by the authors, who take full responsibility for the manuscript.

**Data Availability**
The palaeontological extraction results are publicly accessible through the Treatise.geoLex database at https://treatise.geolex.org/. The data supporting the analyses and Tables 1–3, including the palaeomagnetic and geochemical extraction results used for evaluation, are provided in the accompanying Supplementary Data archive.

**Code Availability**
The source code supporting this study has been provided with the submission for peer review and is currently maintained in a private GitHub repository. It will be made publicly available on GitHub upon publication.

**Funding**
This work was supported by the Key Research Project of Zhejiang Lab (Grant No. 2025SSYS0003).

## References

1. Hey, T., Tansley, S. & Tolle, K. (eds) The Fourth Paradigm: Data-Intensive Scientific Discovery (Microsoft Research, 2009).
2. Bergen, K. J., Johnson, P. A., de Hoop, M. V. & Beroza, G. C. Machine learning for data-driven discovery in solid Earth geoscience. Science **363**, eaau0323 (2019).
3. Wilkinson, M. D. et al. The FAIR Guiding Principles for scientific data management and stewardship. Sci. Data **3**, 160018 (2016).
4. Karpatne, A. et al. Theory-guided data science: A new paradigm for scientific discovery from data. IEEE Trans. Knowl. Data Eng. **29**, 2318–2331 (2017).
5. Crystal-Ornelas, R. et al. Enabling FAIR data in Earth and environmental science with community-centric reporting formats. Sci. Data **9**, 700 (2022).
6. OpenAI. GPT-4 Technical Report. arXiv:2303.08774 (2023).
7. Gemini Team. Gemini 1.5: Unlocking multimodal understanding across millions of tokens of context. arXiv:2403.05530 (2024).
8. Guo, D. et al. DeepSeek-R1: Incentivizing reasoning capability in LLMs via reinforcement learning. arXiv:2501.12948 (2025).
9. Yang, A. et al. Qwen2.5 Technical Report. arXiv:2412.15115 (2024).
10. Yao, S. et al. ReAct: Synergizing reasoning and acting in language models. In International Conference on Learning Representations (2023).
11. Schick, T. et al. Toolformer: Language models can teach themselves to use tools. Adv. Neural Inf. Process. Syst. **36** (2023).
12. Wu, Q. et al. AutoGen: Enabling next-gen LLM applications via multi-agent conversation. arXiv:2308.08155 (2023).
13. Li, G. et al. CAMEL: Communicative agents for ‘mind’ exploration of large language model society. Adv. Neural Inf. Process. Syst. **36** (2023).
14. Ghareeb, A. E. et al. Robin: A multi-agent system for automating scientific discovery. arXiv:2505.13400 (2025).
15. Gottweis, J. et al. Towards an AI co-scientist. arXiv:2502.18864 (2025).
16. Yamada, Y. et al. Towards end-to-end automation of AI research. arXiv:2606.15497 (2026).
17. Aygün, E. et al. An AI system to help scientists write expert-level empirical software. arXiv:2509.06503 (2025).
18. Zhao, W. et al. An agentic system for rare disease diagnosis with traceable reasoning. arXiv:2506.20430 (2025).
19. Anthropic. Introducing the Model Context Protocol. https://www.anthropic.com/news/model-context-protocol (2024). Accessed 10 July 2026.
20. Google Developers Blog. Announcing the Agent2Agent Protocol. https://developers.googleblog.com/en/a2a-a-new-era-of-agent-interoperability/ (2025). Accessed 10 July 2026.
21. Luan, Y., He, L., Ostendorf, M. & Hajishirzi, H. Multi-task identification of entities, relations, and coreference for scientific knowledge graph construction. In Proceedings of the 2018 Conference on Empirical Methods in Natural Language Processing 3219–3232 (2018).
22. Jain, S. et al. SciREX: A challenge dataset for document-level information extraction. In Proceedings of the 58th Annual Meeting of the Association for Computational Linguistics 7506–7516 (2020).
23. Lo, K. et al. S2ORC: The Semantic Scholar Open Research Corpus. In Proceedings of the 58th Annual Meeting of the Association for Computational Linguistics 4969–4983 (2020).

24. Swain, M. C. & Cole, J. M. ChemDataExtractor: A toolkit for automated extraction of chemical information from the scientific literature. J. Chem. Inf. Model. **56**, 1894–1904 (2016).
25. Gupta, T., Zaki, M., Krishnan, N. M. A. & Mausam. MatSciBERT: A materials domain language model for text mining and information extraction. npj Comput. Mater. **8**, 102 (2022).
26. Dagdelen, J. et al. Structured information extraction from scientific text with large language models. Nat. Commun. **15**, 1418 (2024).
27. Polak, M. P. & Morgan, D. Extracting accurate materials data from research papers with conversational language models and prompt engineering. Nat. Commun. **15**, 1569 (2024).
28. Gil, Y. et al. Intelligent systems for geosciences: An essential research agenda. Commun. ACM **62**, 76–84 (2019).
29. Lewis, P. et al. Retrieval-augmented generation for knowledge-intensive NLP tasks. Adv. Neural Inf. Process. Syst. **33**, 9459–9474 (2020).
30. Chen, Y. et al. LongLoRA: Efficient fine-tuning of long-context large language models. In International Conference on Learning Representations (2024).
31. Liu, N. F. et al. Lost in the middle: How language models use long contexts. Trans. Assoc. Comput. Linguist. **12**, 157–173 (2024).
32. Bai, Y. et al. LongBench: A bilingual, multitask benchmark for long context understanding. In Proceedings of the 62nd Annual Meeting of the Association for Computational Linguistics (2024).
33. Cohen-Wang, B., Shah, H., Georgiev, K. et al. ContextCite: Attributing model generation to context. arXiv:2409.00729 (2024).
34. Holzinger, A. Human-in-the-loop machine learning for medical data and knowledge discovery. Data Min. Knowl. Discov. **28**, 1191–1214 (2016).
35. Schleith, J., Hoffmann, H., Norkute, M. & Cechmanek, B. Human-in-the-loop information extraction increases efficiency and trust. In Mensch und Computer 2022 – Workshopband (2022).
36. GeoGPT-Research-Project. GeoGPT-R1-Preview and Qwen2.5-72B-GeoGPT. Hugging Face model cards. https://huggingface.co/GeoGPT-Research-Project (2026). Accessed 10 July 2026.
37. Moore, R. C., Teichert, C. & subsequent editors. *Treatise on Invertebrate Paleontology* (Geological Society of America, University of Kansas Press and Paleontological Institute, 1953–present).
38. Peters, S. E. & McClennen, M. The Paleobiology Database application programming interface. Paleobiology **42**, 1–7 (2016).
39. International Commission on Zoological Nomenclature. International Code of Zoological Nomenclature 4th edn (International Trust for Zoological Nomenclature, 1999).
40. Cohen, K. M., Finney, S. C., Gibbard, P. L. & Fan, J.-X. The ICS International Chronostratigraphic Chart. Episodes **36**, 199–204 (2013).
41. GeoLex. *Treatise on Invertebrate Paleontology* Fossil Search. https://treatise.geolex.org/ (2026). Accessed 10 July 2026.
42. Tauxe, L. et al. PmagPy: Software package for palaeomagnetic data analysis and a bridge to the Magnetics Information Consortium (MagIC) database. Geochem. Geophys. Geosyst. **17**, 2450–2463 (2016).
43. GEOROC. Geochemistry of Rocks of the Oceans and Continents database. https://georoc.eu/ (2026). Accessed 10 July 2026.
44. Lehnert, K., Su, Y., Langmuir, C. H., Sarbas, B. & Nohl, U. A global geochemical database structure for rocks. Geochem. Geophys. Geosyst. **1**, 1012 (2000).

45. Ball, P. W. Global distribution and composition of Neogene–Quaternary intraplate volcanic rocks. GFZ Data Services. https://doi.org/10.5880/digis.e.2024.006 (2024).
46. Binmakhashen, G. M. & Mahmoud, S. A. Document layout analysis: A comprehensive survey. ACM Comput. Surv. **52**, 109 (2019).
47. Smock, B., Pesala, R. & Abraham, R. PubTables-1M: Towards comprehensive table extraction from unstructured documents. In Proceedings of the IEEE/CVF Conference on Computer Vision and Pattern Recognition 4634–4642 (2022).
48. Ling, J. et al. Table2LaTeX-RL: High-fidelity LaTeX code generation from table images via reinforced multimodal language models. arXiv:2509.17589 (2025).
49. Bazzo, G. T., Lorentz, G. A., Vargas, D. S. & Moreira, V. P. Assessing the impact of OCR errors in information retrieval. In Advances in Information Retrieval, ECIR 2020 102–109 (2020).
50. Mathpix, Inc. Mathpix: AI-powered document conversion for STEM. https://mathpix.com/ (2026). Accessed 10 July 2026.
51. Liu, J. LlamaIndex. https://github.com/run-llama/llama_index (2022). Accessed 10 July 2026.
52. Xiao, S., Liu, Z., Zhang, P. & Muennighoff, N. C-Pack: Packaged resources to advance general Chinese embedding. arXiv:2309.07597 (2023).
53. Robertson, S. & Zaragoza, H. The probabilistic relevance framework: BM25 and beyond. Found. Trends Inf. Retr. **3**, 333–389 (2009).
54. NetEase Youdao. bce-reranker-base_v1. Hugging Face model card. https://huggingface.co/maidalun1020/bce-reranker-base_v1 (2026). Accessed 10 July 2026.
55. Mishra, S., Mendhe, V. A., Varma, A. K., Kamble, A. D. et al. Influence of organic and inorganic content on fractal dimensions of Barakar and Barren Measures shale gas reservoirs of Raniganj basin, India. J. Nat. Gas Sci. Eng. **49**, 393–409 (2018).
56. Chen, J. et al. M3-embedding: Multi-linguality, multi-functionality, multi-granularity text embeddings through self-knowledge distillation. In Findings of the Association for Computational Linguistics: ACL 2024 2318–2335 (2024).
57. BAAI. bge-reranker-large. Hugging Face model card. https://huggingface.co/BAAI/bge-reranker-large (2026). Accessed 10 July 2026.
58. DeepSeek-AI et al. DeepSeek-V3 Technical Report. arXiv:2412.19437 (2024).
59. Smith, J. A. et al. Increasing the equitability of data citation in paleontology: capacity building for the big data future. *Paleobiology* **50**, 165–176 (2024).

**Acknowledgements**

We thank Aditya Sivathanu and Kevin Chang for their contributions to the website presentation and online display of the extracted Treatise data. We also thank the GeoGPT team and collaborators involved in system testing, data organization, and technical discussions. We are grateful to the contributors and editors of the *Treatise on Invertebrate Paleontology*, whose long-term scholarly work provided the foundation for this study.

**Author Contributions**

Ziqi Song, Zongyuan Xiang, and Jiang Yang conceived the study. Ziqi Song and Zongyuan Xiang designed and developed the HERMES framework. Zongyuan Xiang implemented the extraction pipeline, performed the experiments, conducted the large-scale data extraction, and analyzed the results. Ziqi Song, Yufei Ye, Shuan Li, Zhong Peng, and Shaoqi Yu contributed to framework implementation, methodological development, and result analysis. James G. Ogg, Gabi Ogg, Wen Du, and Juye Wei contributed to data curation, data validation, and quality assessment of the extracted records. Bruce S. Lieberman and Natalia López Carranza contributed to the preparation and organization of the digitized *Treatise on Invertebrate Paleontology* materials, helped frame the study within the FAIR data context, and provided comparative test datasets. James G. Ogg, Bruce S. Lieberman, and Natalia López Carranza provided palaeontological expertise, identified recurring structural patterns and shared elements in the Treatise text, and guided the interpretation of the Treatise data. Zongyuan Xiang and Ziqi Song prepared the figures and tables and wrote the original draft of the manuscript. James G. Ogg and Bruce S. Lieberman contributed to reviewing and editing the manuscript. Jiang Yang and Ying Zhou supervised the work. Jiang Yang, Ying Zhou, and Jieping Ye contributed to project administration, with Jiang Yang also providing conceptual and strategic guidance. Ying Zhou and Jieping Ye acquired and managed funding for the project. All authors reviewed and approved the final manuscript.

**Competing Interests**

The authors declare no competing interests.

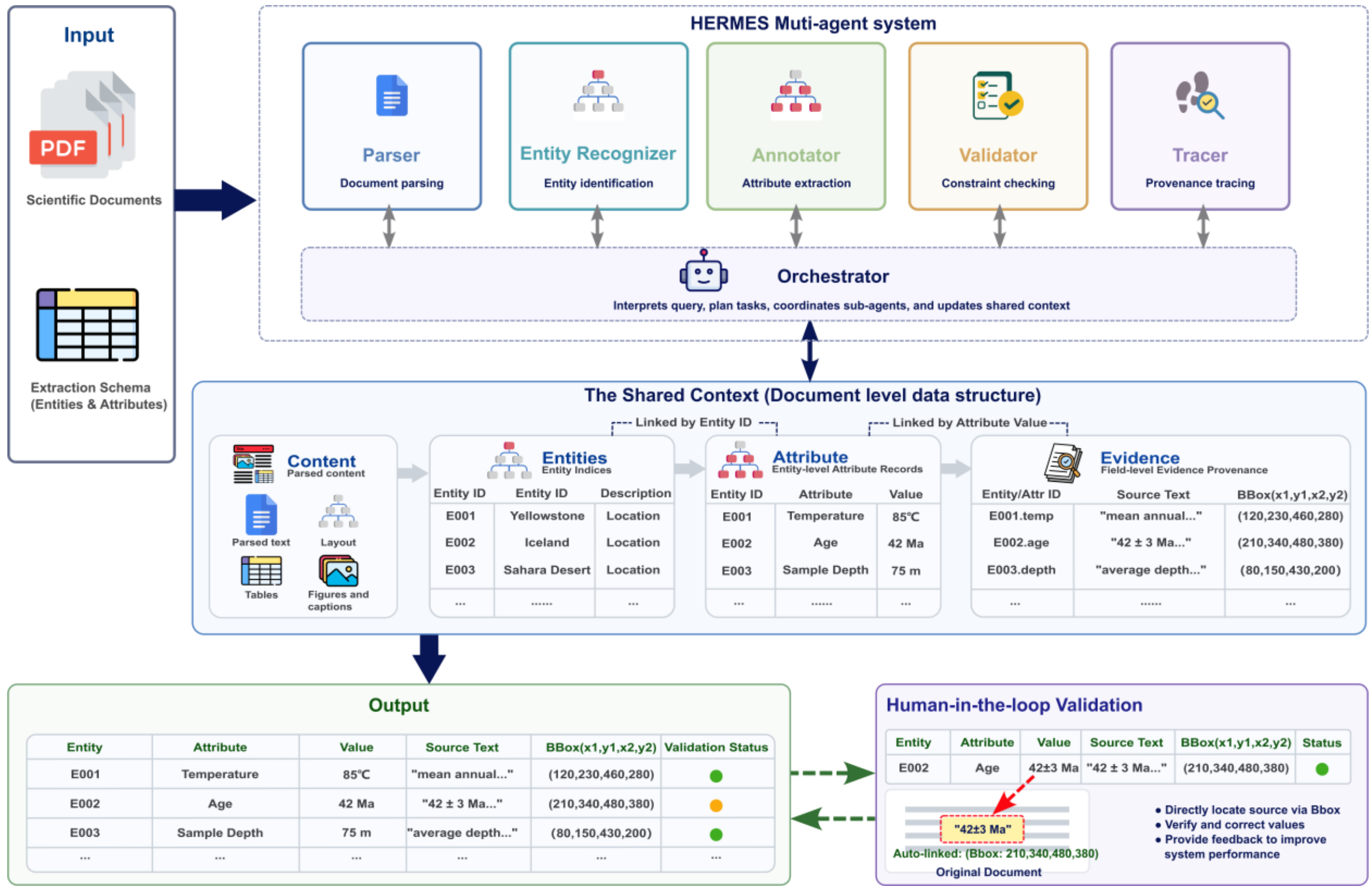

**Fig. 1. Overall architecture of HERMES.** The Orchestrator coordinates the Parser, Entity Recognizer, Annotator, Validator, and Tracer through a shared document-level context, producing structured entity–attribute outputs linked to source evidence and supporting human-in-the-loop validation.

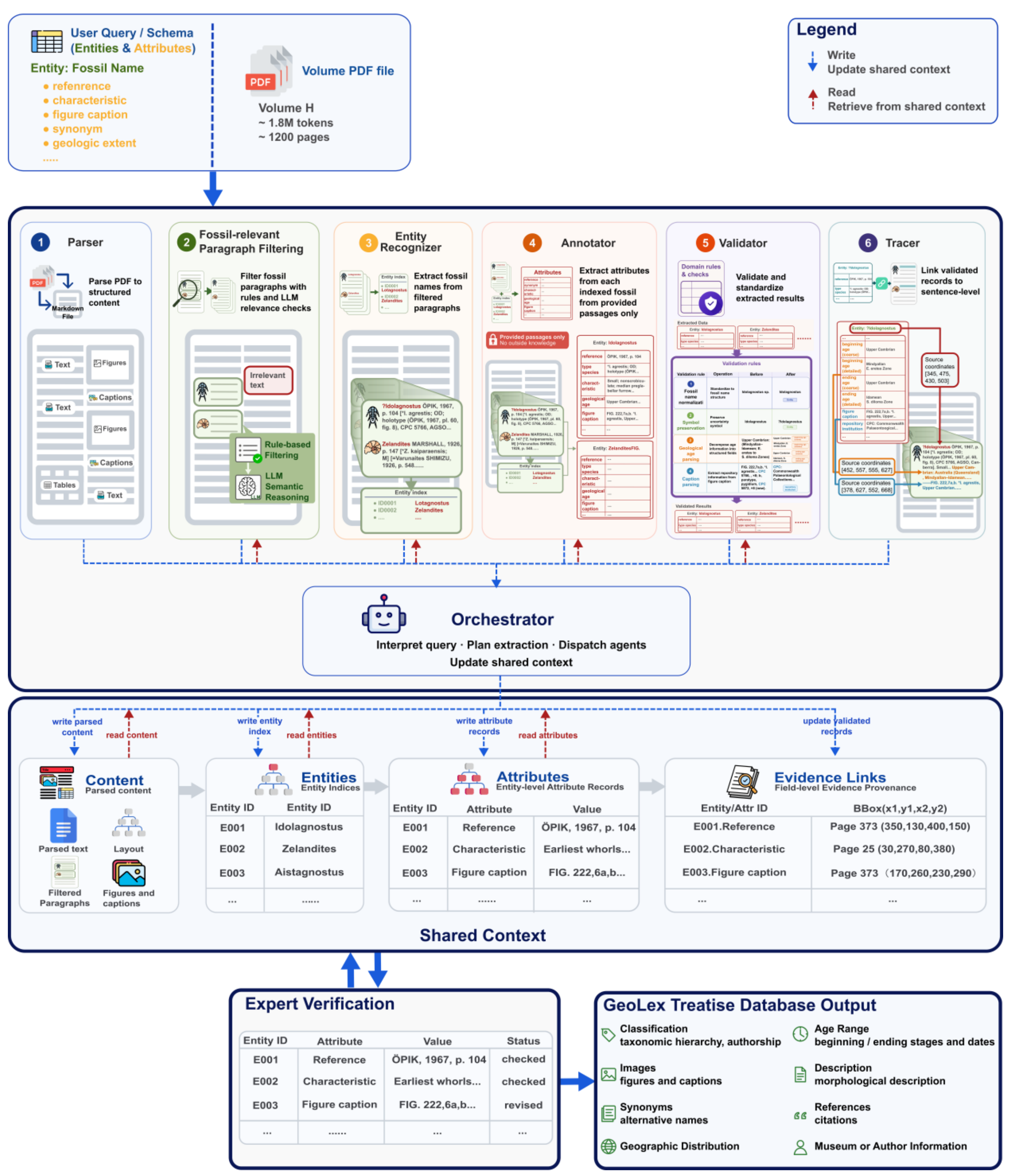

| Entity ID | Entity ID |
|---|---|
| E001 | Idolagnostus |
| E002 | Zelandites |
| E003 | Aistagnostus |
| ... | ...... |

| Entity ID | Attribute | Value |
|---|---|---|
| E001 | Reference | ÖPIK, 1967, p. 104 |
| E002 | Characteristic | Earliest whorls... |
| E003 | Figure caption | FIG. 222,6a,b... |
| ... | ...... | ... |

| Entity/Attr ID | BBox(x1,y1,x2,y2) |
|---|---|
| E001.Reference | Page 373 (350,130,400,150) |
| E002.Characteristic | Page 25 (30,270,80,380) |
| E003.Figure caption | Page 373 (170,260,230,290) |
| ... | ... |

| Entity ID | Attribute | Value | Status |
|---|---|---|---|
| E001 | Reference | ÖPIK, 1967, p. 104 | checked |
| E002 | Characteristic | Earliest whorls... | checked |
| E003 | Figure caption | FIG. 222,6a,b... | revised |
| ... | ...... | ... | ... |

**Fig. 2. Fossil data extraction process from the *Treatise on Invertebrate Paleontology*.** HERMES identifies fossil-relevant passages from each Treatise volume, extracts fossil entities and their attributes, applies domain-specific validation rules, and links extracted records to source evidence for expert verification.

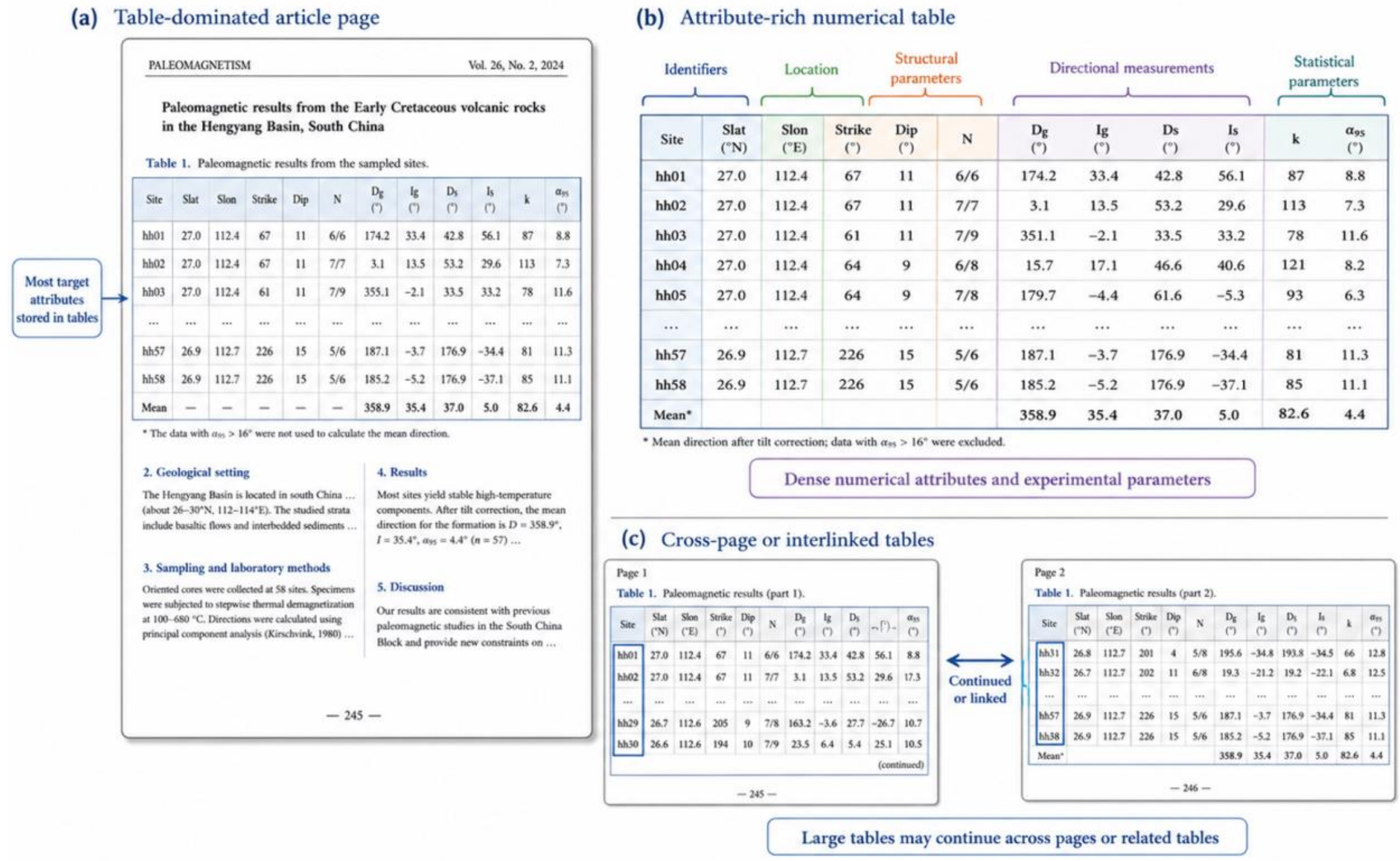

PALEOMAGNETISM Vol. 26, No. 2, 2024

**Paleomagnetic results from the Early Cretaceous volcanic rocks in the Hengyang Basin, South China**

**Table 1.** Paleomagnetic results from the sampled sites.

| Site | Slat | Slon | Strike | Dip | N | Dg (°) | Ig (°) | Ds (°) | Is (°) | k | $\alpha_{95}$ (°) |
|---|---|---|---|---|---|---|---|---|---|---|---|
| hh01 | 27.0 | 112.4 | 67 | 11 | 6/6 | 174.2 | 33.4 | 42.8 | 56.1 | 87 | 8.8 |
| hh02 | 27.0 | 112.4 | 67 | 11 | 7/7 | 3.1 | 13.5 | 53.2 | 29.6 | 113 | 7.3 |
| hh03 | 27.0 | 112.4 | 61 | 11 | 7/9 | 355.1 | -2.1 | 33.5 | 33.2 | 78 | 11.6 |
| ... | ... | ... | ... | ... | ... | ... | ... | ... | ... | ... | ... |
| hh57 | 26.9 | 112.7 | 226 | 15 | 5/6 | 187.1 | -3.7 | 176.9 | -34.4 | 81 | 11.3 |
| hh58 | 26.9 | 112.7 | 226 | 15 | 5/6 | 185.2 | -5.2 | 176.9 | -37.1 | 85 | 11.1 |
| Mean | — | — | — | — | — | 358.9 | 35.4 | 37.0 | 5.0 | 82.6 | 4.4 |

* The data with $\alpha_{95}$ > 16° were not used to calculate the mean direction.

**2. Geological setting**

The Hengyang Basin is located in south China ... (about 26–30°N, 112–114°E). The studied strata include basaltic flows and interbedded sediments ...

**3. Sampling and laboratory methods**

Oriented cores were collected at 58 sites. Specimens were subjected to stepwise thermal demagnetization at 100–680 °C. Directions were calculated using principal component analysis (Kirschvink, 1980) ...

**4. Results**

Most sites yield stable high-temperature components. After tilt correction, the mean direction for the formation is $D = 358.9°$, $I = 35.4°$, $\alpha_{95} = 4.4°$ ($n = 57$) ...

**5. Discussion**

Our results are consistent with previous paleomagnetic studies in the South China Block and provide new constraints on ...

— 245 —

| Site | Slat (°N) | Slon (°E) | Strike (°) | Dip (°) | N | Dg (°) | Ig (°) | Ds (°) | Is (°) | k | $\alpha_{95}$ (°) |
|---|---|---|---|---|---|---|---|---|---|---|---|
| hh01 | 27.0 | 112.4 | 67 | 11 | 6/6 | 174.2 | 33.4 | 42.8 | 56.1 | 87 | 8.8 |
| hh02 | 27.0 | 112.4 | 67 | 11 | 7/7 | 3.1 | 13.5 | 53.2 | 29.6 | 113 | 7.3 |
| hh03 | 27.0 | 112.4 | 61 | 11 | 7/9 | 351.1 | −2.1 | 33.5 | 33.2 | 78 | 11.6 |
| hh04 | 27.0 | 112.4 | 64 | 9 | 6/8 | 15.7 | 17.1 | 46.6 | 40.6 | 121 | 8.2 |
| hh05 | 27.0 | 112.4 | 64 | 9 | 7/8 | 179.7 | −4.4 | 61.6 | −5.3 | 93 | 6.3 |
| ... | ... | ... | ... | ... | ... | ... | ... | ... | ... | ... | ... |
| hh57 | 26.9 | 112.7 | 226 | 15 | 5/6 | 187.1 | −3.7 | 176.9 | −34.4 | 81 | 11.3 |
| hh58 | 26.9 | 112.7 | 226 | 15 | 5/6 | 185.2 | −5.2 | 176.9 | −37.1 | 85 | 11.1 |
| Mean* | | | | | | 358.9 | 35.4 | 37.0 | 5.0 | 82.6 | 4.4 |

* Mean direction after tilt correction; data with $\alpha_{95}$ > 16° were excluded.

Page 1

**Table 1.** Paleomagnetic results (part 1).

| Site | Slat (°N) | Slon (°E) | Strike (°) | Dip (°) | N | Dg (°) | Ig (°) | Ds (°) | –(°)– | $\alpha_{95}$ (°) |
|---|---|---|---|---|---|---|---|---|---|---|
| hh01 | 27.0 | 112.4 | 67 | 11 | 6/6 | 174.2 | 33.4 | 42.8 | 56.1 | 8.8 |
| hh02 | 27.0 | 112.4 | 67 | 11 | 7/7 | 3.1 | 13.5 | 53.2 | 29.6 | 17.3 |
| ... | ... | ... | ... | ... | ... | ... | ... | ... | ... | ... |
| hh29 | 26.7 | 112.6 | 205 | 9 | 7/8 | 163.2 | -3.6 | 27.7 | -26.7 | 10.7 |
| hh30 | 26.6 | 112.6 | 194 | 10 | 7/9 | 23.5 | 6.4 | 5.4 | 25.1 | 10.5 |

(continued)

— 245 —

Page 2

**Table 1.** Paleomagnetic results (part 2).

| Site | Slat (°N) | Slon (°E) | Strike (°) | Dip (°) | N | Dg (°) | Ig (°) | Ds (°) | Is (°) | k | $\alpha_{95}$ (°) |
|---|---|---|---|---|---|---|---|---|---|---|---|
| hh31 | 26.8 | 112.7 | 201 | 4 | 5/8 | 195.6 | −34.8 | 193.8 | −34.5 | 66 | 12.8 |
| hh32 | 26.7 | 112.7 | 202 | 11 | 6/8 | 19.3 | −21.2 | 19.2 | −22.1 | 6.8 | 12.5 |
| ... | ... | ... | ... | ... | ... | ... | ... | ... | ... | ... | ... |
| hh57 | 26.9 | 112.7 | 226 | 15 | 5/6 | 187.1 | −3.7 | 176.9 | −34.4 | 81 | 11.3 |
| hh38 | 26.9 | 112.7 | 226 | 15 | 5/6 | 185.2 | −5.2 | 176.9 | −37.1 | 85 | 11.1 |
| Mean* | | | | | | 358.9 | 35.4 | 37.0 | 5.0 | 82.6 | 4.4 |

— 246 —

**Fig. 3. Representative palaeomagnetic documents and extraction challenges.** Palaeomagnetic literature commonly contains site- and sample-level measurements in dense numerical tables, requiring table-aware extraction and reliable binding of directional, statistical, age, and locality attributes to the correct records.

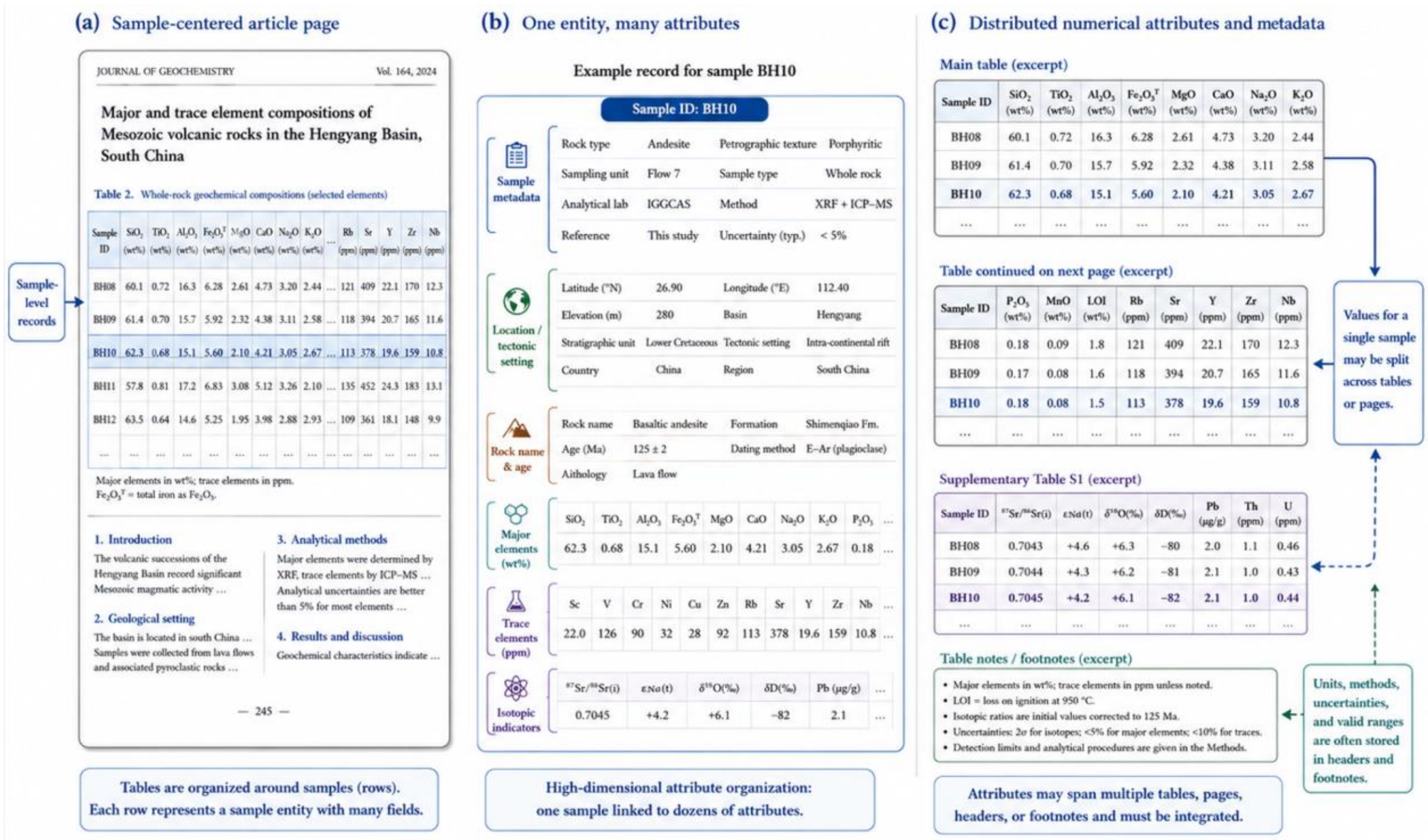

**Fig. 4. Representative geochemical documents and extraction challenges.** Geochemical literature is organized around sample-level entities associated with high-dimensional attributes, including location, age, rock type, major elements, trace elements, isotopic ratios, units, and analytical metadata.

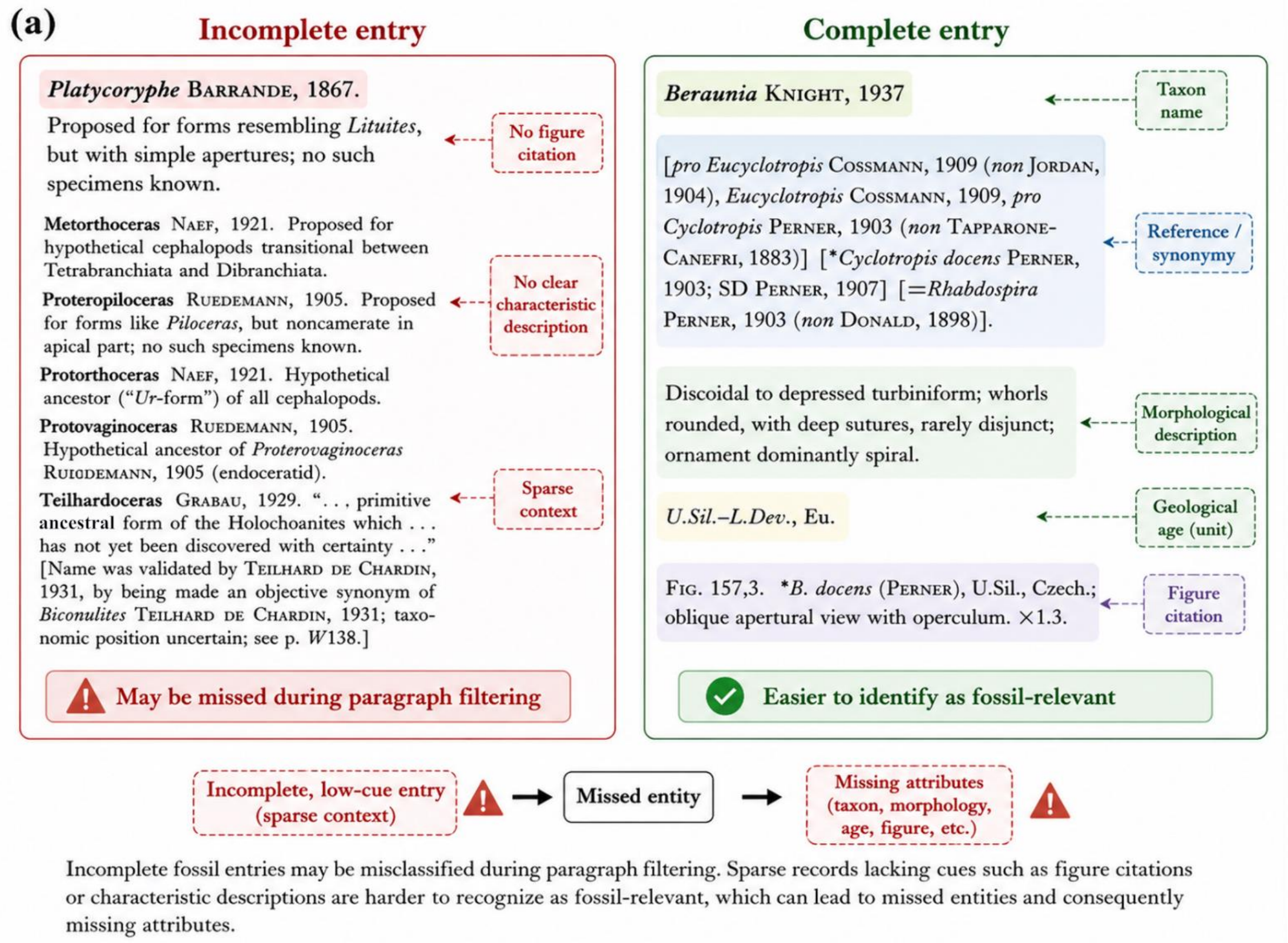

(a) Incomplete fossil entries may be misclassified during paragraph filtering: the example on the left lacks sufficient descriptive context for HERMES, whereas the example on the right contains sufficient taxonomic and morphologic information.

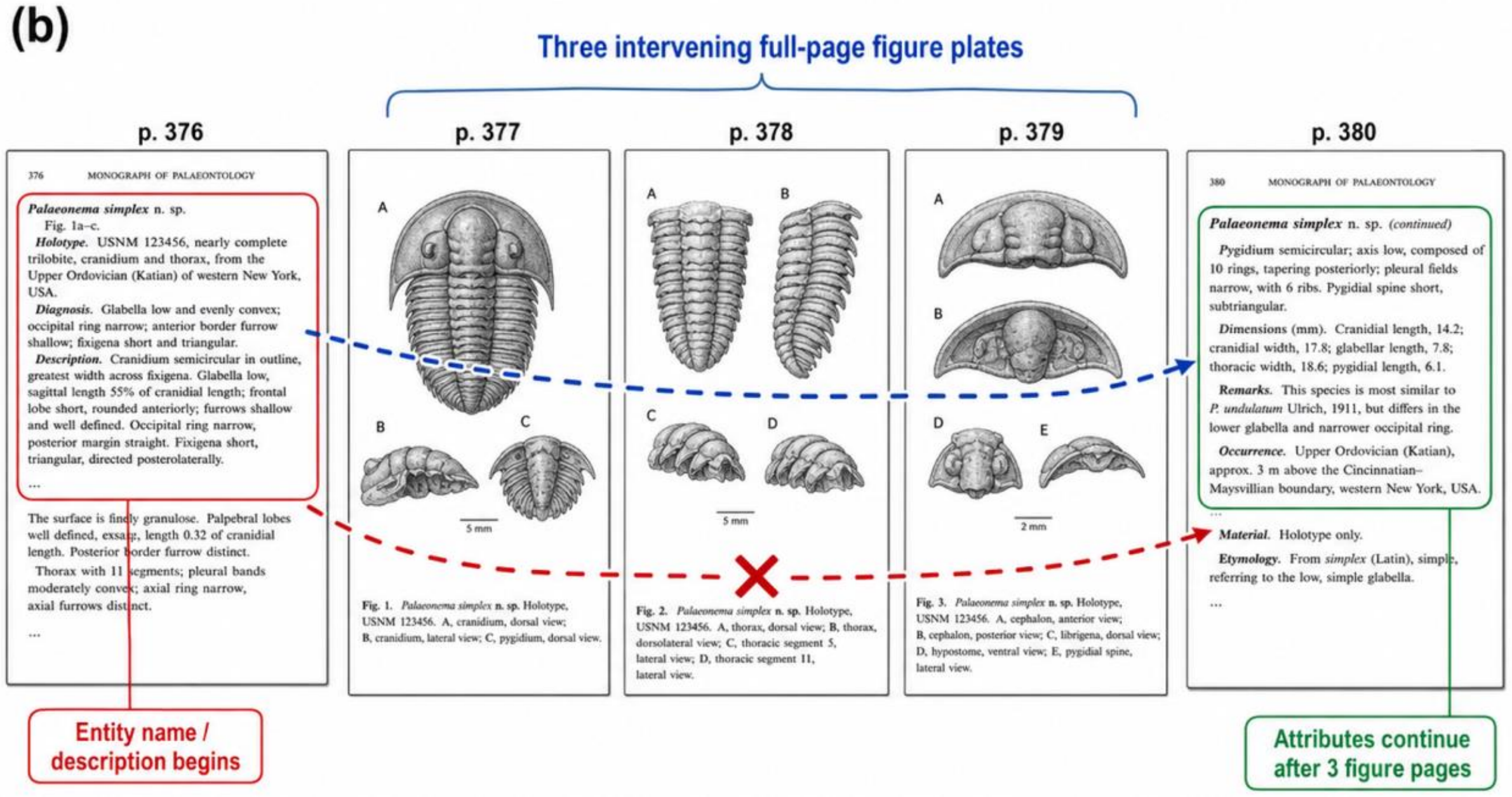

(b) Cross-page discontinuity can disrupt parsing by HERMES: fossil descriptions may be split across page boundaries, and intervening figures or captions can cause the preceding and following text fragments to be misaligned or incorrectly separated.

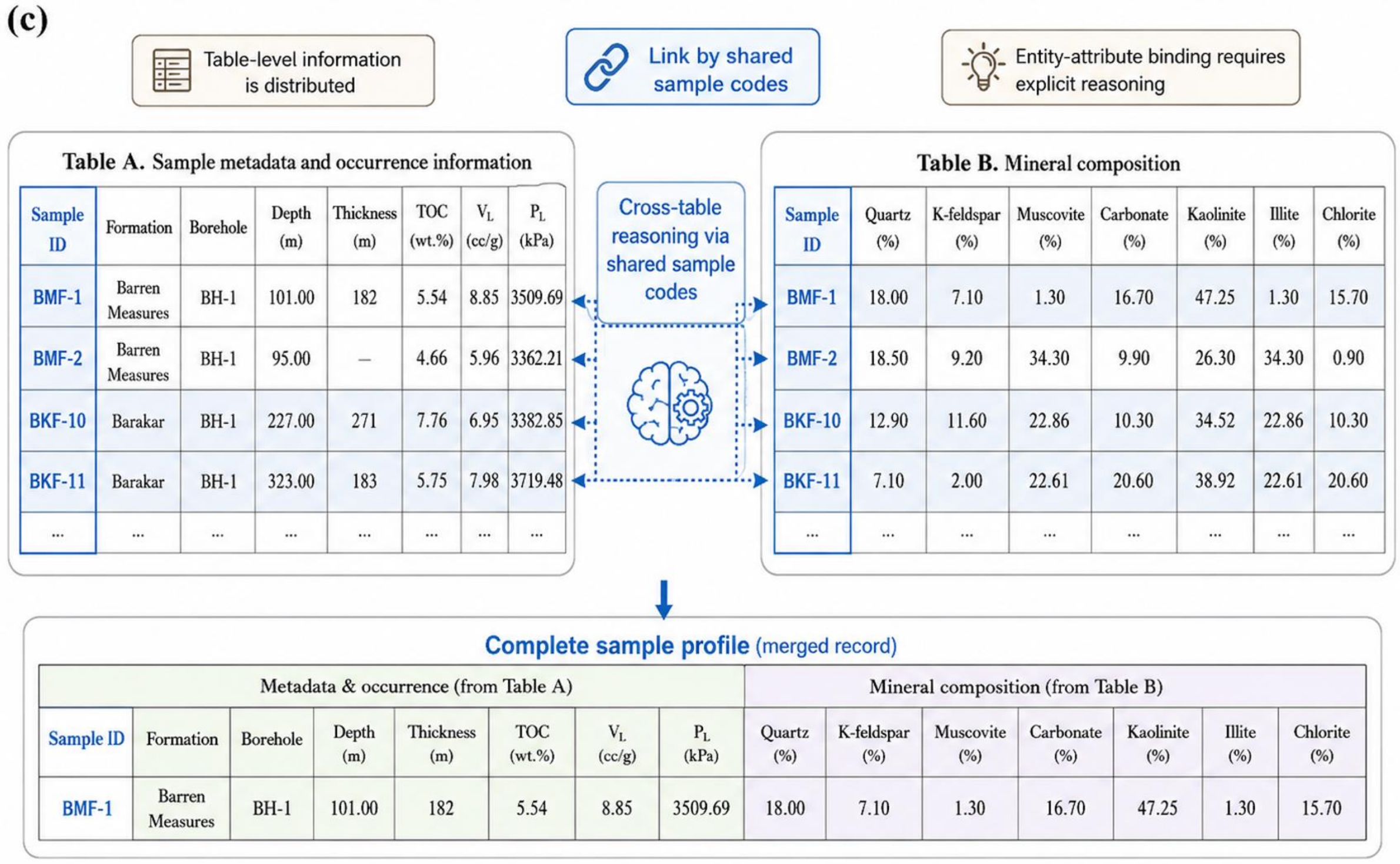

(c) In some geoscience literature, sample-level information is distributed across multiple tables, requiring cross-table reasoning. For example, one table records sample IDs together with formation, borehole, depth, thickness, total organic carbon (TOC), and sorption-related properties, while another table reports mineral compositions for the same sample IDs. Extracting complete sample profiles therefore requires linking records across tables by shared sample codes.

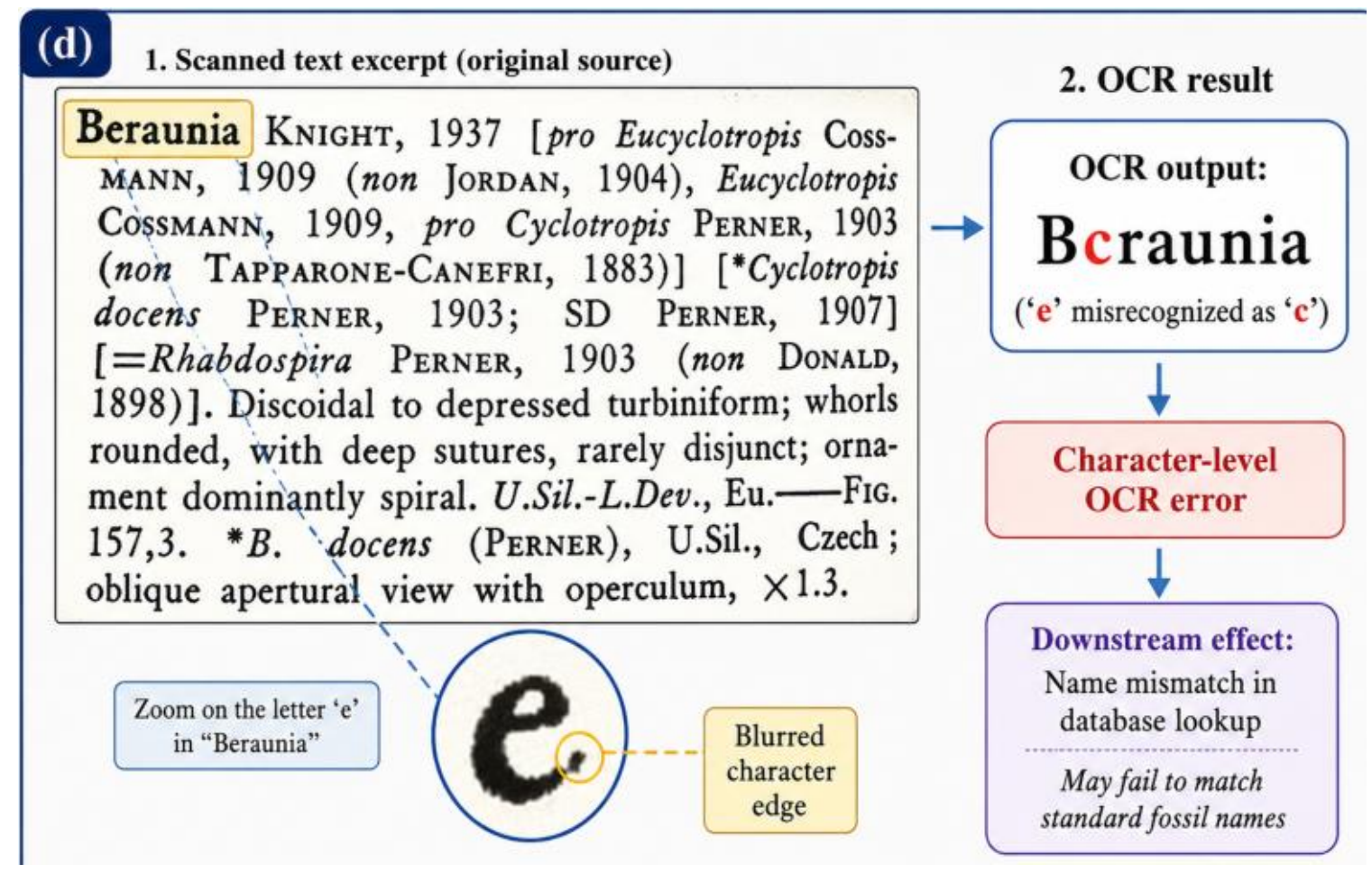

(d) The fossil name "*Beraunia*" may be recognized as "*Bcraunia*" due to blurred edges of the letter 'e'.

**Fig. 5. Representative error sources in automated extraction from complex scientific documents.** Examples include incomplete fossil entries, cross-page discontinuity, cross-table sample linkage, and OCR-induced name errors, all of which can affect entity recognition, attribute binding, or evidence tracing.

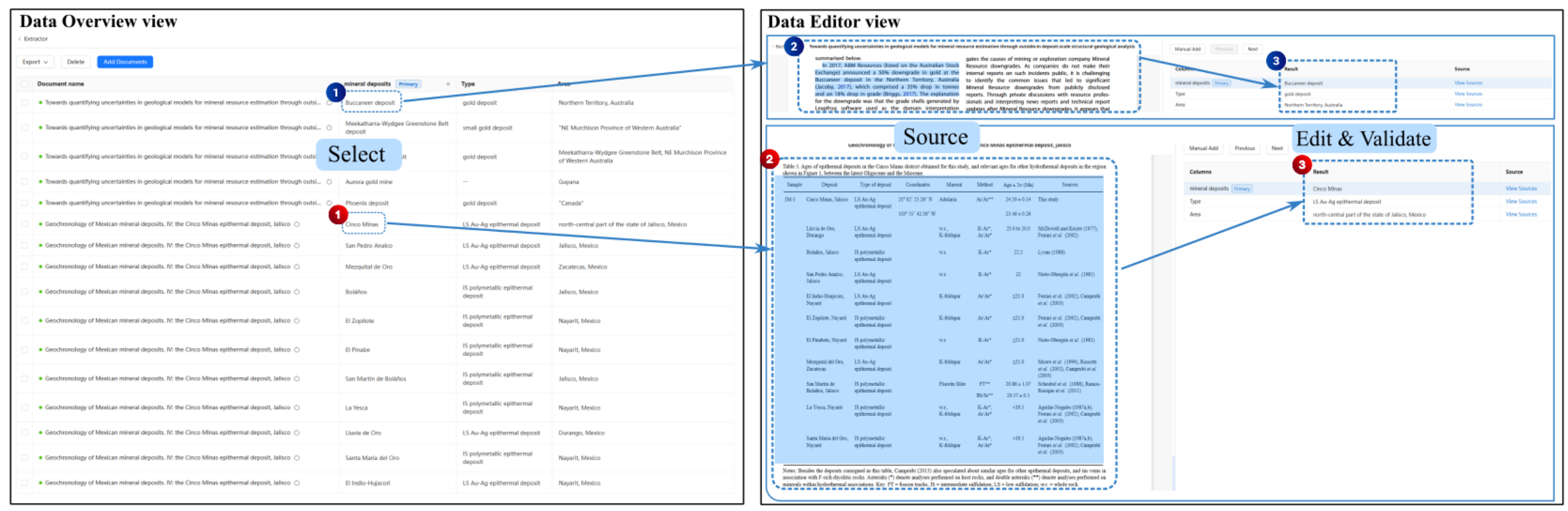

**Fig. 6. Traceable evidence-based validation and editing of extracted data.** The validation interface links each extracted entity or attribute value to its source evidence in the original document, allowing domain experts to inspect, edit, and confirm structured outputs in context.

| Fossil group | Entities, n | Entity | | | Attribute | | |
|---|---|---|---|---|---|---|---|
| | | Recall | Precision | F1 Score | Recall | Precision | F1 score |
| Graptolites | 482 | 0.94 | 0.97 | 0.95 | 0.92 | 0.96 | 0.94 |
| Charophyta | 96 | 0.90 | 0.98 | 0.93 | 0.90 | 0.99 | 0.94 |
| Brachiopods | 4,676 | 0.94 | 0.99 | 0.97 | 0.93 | 0.99 | 0.96 |
| Trilobites | 1,791 | 0.80 | 0.87 | 0.84 | 0.83 | 0.98 | 0.89 |
| Ammonoids | 2,456 | 0.92 | 0.75 | 0.83 | 0.89 | 0.98 | 0.93 |
| Porifera | 2,208 | 0.95 | 0.96 | 0.95 | 0.94 | 0.96 | 0.95 |
| Crinoids | 1,123 | 0.94 | 0.98 | 0.96 | 0.93 | 0.98 | 0.96 |
| Bryozoa | 1,410 | 0.85 | 0.98 | 0.91 | 0.87 | 0.85 | 0.86 |
| Molluscs | 2,366 | 0.93 | 0.96 | 0.94 | 0.90 | 0.96 | 0.92 |
| Arthropoda | 1,441 | 0.88 | 0.93 | 0.90 | 0.86 | 0.92 | 0.89 |
| Conodonts | 243 | 0.96 | 0.92 | 0.94 | 0.89 | 0.97 | 0.93 |
| Protista | 3,334 | 0.76 | 0.95 | 0.85 | 0.73 | 0.81 | 0.77 |
| Bivalves | 6,934 | 0.83 | 0.89 | 0.86 | 0.80 | 0.95 | 0.87 |
| Corals | 997 | 0.89 | 0.60 | 0.72 | 0.84 | 0.92 | 0.88 |
| Echinoderms | 2,011 | 0.95 | 0.79 | 0.86 | 0.76 | 0.95 | 0.84 |
| Trace Fossils | 709 | 0.98 | 0.88 | 0.92 | 0.88 | 0.95 | 0.91 |
| Overall | 32,277 | 0.89 | 0.90 | 0.90 | 0.87 | 0.95 | 0.91 |

**Table 1. HERMES performance across major fossil groups in the *Treatise on Invertebrate Paleontology*.** The original, unedited HERMES outputs were compared with an expert-adjudicated reference dataset produced through complete review of the original PDF volumes. During this review, experts corrected erroneous records, removed spurious records, and added entities and attributes missed by the system. All records from the evaluated volumes were included, without subsampling. *Entities, n* denotes the number of entities in each fossil group. Each entity was evaluated against the same set of 14 predefined attributes; therefore, the number of evaluated attributes for each group was $14 \times n$ and is not separately reported. An empty attribute indicates that the corresponding information was not reported in the source document. Precision, recall and F1 score are reported at the entity and attribute levels.

**Definitions used in Tables 1–3:**

True positives (TP): correctly extracted reference records

False positives (FP): extracted records that were incorrect or absent from the reference dataset

False negatives (FN): reference records not extracted by the system

F1 score is the harmonic mean of recall and precision.

$$\text{Recall} = \frac{TP}{TP + FN}$$

$$\text{Precision} = \frac{TP}{TP + FP}$$

$$\text{F1} = 2 \times \frac{Precision \times Recall}{Precision + Recall}$$

| References | Entities, n | Entity | | | Attribute | | |
|---|---|---|---|---|---|---|---|
| | | Recall | Precision | F1 Score | Recall | Precision | F1 score |
| Wang et al. (2016) | 21 | 1.00 | 1.00 | 1.00 | 0.77 | 0.88 | 0.82 |
| Funahara et al. (1992) | 30 | 1.00 | 1.00 | 1.00 | 0.84 | 0.84 | 0.84 |
| Li et al. (2013) | 13 | 1.00 | 1.00 | 1.00 | 0.99 | 0.87 | 0.92 |
| Otofuji et al. (1998) | 21 | 1.00 | 1.00 | 1.00 | 0.86 | 0.76 | 0.81 |
| Sun et al. (2006) | 57 | 0.91 | 0.91 | 0.91 | 0.76 | 0.87 | 0.81 |
| Huang et al. (2013) | 26 | 0.73 | 0.91 | 0.81 | 0.55 | 0.77 | 0.64 |
| Gilder et al. (1993) | 12 | 1.00 | 0.80 | 0.89 | 0.30 | 0.40 | 0.34 |
| Zhai et al. (1992) | 10 | 1.00 | 1.00 | 1.00 | 0.83 | 0.63 | 0.72 |
| Narumoto et al. (2006) | 43 | 1.00 | 1.00 | 1.00 | 0.87 | 0.84 | 0.86 |
| Li et al. (2005) | 12 | 0.92 | 0.48 | 0.63 | 0.74 | 0.80 | 0.77 |
| Liu et al. (1999) | 18 | 1.00 | 1.00 | 1.00 | 0.86 | 0.77 | 0.81 |
| Zhu et al. (2006) | 30 | 1.00 | 1.00 | 1.00 | 0.61 | 0.79 | 0.69 |
| Gao et al. (2017) | 23 | 1.00 | 1.00 | 1.00 | 0.99 | 0.94 | 0.96 |
| Overall | 316 | 0.96 | 0.93 | 0.94 | 0.77 | 0.81 | 0.79 |

**Table 2. HERMES performance on palaeomagnetic literature.** Each row represents one evaluated palaeomagnetic source document, identified by the first author and publication year. *Entities, n* denotes the number of entities in each document. Each entity was evaluated against the same set of 10 predefined attributes; therefore, the number of evaluated attributes for each document was $10 \times n$ and is not reported separately. Entity- and attribute-level recall, precision, and F1 score are reported for each document and for the overall evaluation set.

| References | Entities, n | Entity | | | Attribute | | |
|---|---|---|---|---|---|---|---|
| | | Recall | Precision | F1 Score | Recall | Precision | F1 score |
| Sheth et al. (2003) | 14 | 1.00 | 1.00 | 1.00 | 0.38 | 1.00 | 0.55 |
| Ho et al. (2008) | 27 | 0.52 | 0.37 | 0.43 | 0.27 | 0.93 | 0.42 |
| Choi et al. (2006) | 8 | 0.88 | 0.54 | 0.67 | 0.38 | 0.91 | 0.53 |
| Kamgang et al. (2013) | 33 | 1.00 | 0.94 | 0.97 | 0.32 | 0.93 | 0.47 |
| Yahiaoui et al. (2014) | 6 | 1.00 | 0.86 | 0.92 | 0.75 | 1.00 | 0.86 |
| Paul et al. (2007) | 31 | 1.00 | 0.86 | 0.93 | 0.36 | 1.00 | 0.53 |
| Taneja et al. (2016) | 8 | 1.00 | 0.89 | 0.94 | 0.27 | 0.57 | 0.37 |
| Overall | 127 | 0.89 | 0.74 | 0.81 | 0.35 | 0.93 | 0.51 |

**Table 3. HERMES performance on geochemical literature.** Each row represents one evaluated geochemical source document, identified by the first author and publication year. *Entities, n* denotes the number of entities in each document. The Overall row includes 127 reference entities. Each entity was evaluated against the same set of 10 predefined attributes; therefore, the number of evaluated attributes for each document was $10 \times n$ and is not reported separately. Entity- and attribute-level recall, precision, and F1 score are reported for each document and for the overall evaluation set.

Supplementary Fig. 1

Paleontological extraction query.

**Paleontological fossil extraction query**

**Entity Recognition Prompt**

**System:**
Extract all <MainKey> from the given document.
Only include <MainKey> that are explicitly mentioned in the document.
If no <MainKey> are found, respond with:
"No <MainKey> found in the document."
For each <MainKey>, provide:
<MainKey>: [entity name or identifier]
Description: [brief description or value]

Do not list results with indices.

Context information is below.
**{context_str}**

**User:**
Extract the following entity from the document.
MainKey Name: Formal genus name
Data Type: string
Description: The formal genus name. MUST be the first single-word fossil name of each given paragraph. For subgenera indicated in parentheses like P. (Patella), extract the FULL name including both the abbreviated genus and the subgenus in parentheses. Do not omit the parentheses or any part of the subgenus notation.
Example: Lingula

**Structured Output Format:**

[DATA_ITEM]
Key:
Lingula
[SUB_ITEMS]
SubKey: Formal genus name and reference
SubType: string
SubValues: Lingula BRUGUIÈRE, 1797, pl. 250
SubSource: Genus description

SubKey: Type species
SubType: string
SubValues: [*L. anatine LAMARCK, 1801...... 1985]
SubSource: First square bracket after genus name

SubKey: Synonyms
SubType: string
SubValues: [=Pharetra BOLTEN, 1798, ......, p. 76]
SubSource: Synonym bracket beginning with "="

SubKey: Distinguishing characteristics
SubType: string
SubValues: Shell elongate-oval ...... morphology.
SubSource: Descriptive diagnosis paragraph

SubKey: Geological periods
SubType: string
SubValues: Holocene
SubSource: Genus occurrence statement / figure caption

SubKey: Geographic extent
SubType: string
SubValues: Australia; Fiji
SubSource: Occurrence statement / figure caption

SubKey: Figure captions
SubType: string
SubValues: ——FIG. 8,1a–f. *L. anatine ...... ×1.9 (new).
SubSource: Figure caption

[END_DATA_ITEM]
[END_OF_EXTRACTION]

**Attribute Extraction Prompt**

**System:**
Extract the following targeted data from the given context with high precision.
Please provide an answer based solely on the provided sources.
When referencing information from a source, cite the appropriate source(s) using their corresponding numbers.
Every answer should include at least one source citation.
Only cite a source when you are explicitly referencing it.
If none of the sources are helpful, you should indicate that.
Instructions for precise extraction:
1.For numeric values, always provide the exact numbers found in the text. Do not summarize or provide ranges unless explicitly stated in the source.
2.If a range is given in the source, provide both the lower and upper bounds exactly as stated.
3.Include units of measurement when present.
4.For non-numeric data, provide the full, exact text as it appears in the source.
5.If the information is not available or unclear, state "Information not available" instead of leaving it blank or providing a partial answer.
6.Do not repeat the same value if the same value is found in multiple sources.

Context information is below.
**{context_str}**

Given the context information and not prior knowledge, answer the query using the specified structured output format.
Create a separate [DATA_ITEM] block for each item extracted from the context.
You can repeat the [DATA_ITEM] ... [END_DATA_ITEM] structure for multiple items.
Inside each [DATA_ITEM], you can have multiple sub-items in the [SUB_ITEMS] section.

After processing all data items, end your response with:
[END_OF_EXTRACTION]

User:
For the [main key] 'Lingula', please list the following sub-items:
- Formal genus name and reference (string), refers to the formal scientific name of a genus, followed by the author, year, and page or plate reference. Extract all information before the first set of square brackets.
- Type species (string), refers to the type species indicated by the first set of square brackets, usually beginning with an asterisk. Extract the full content inside the corresponding square brackets, including abbreviations such as "OD."
- Synonyms (string), refers to alternative genus names listed in square brackets that begin with an equal sign (=). Extract all such bracketed synonym contents completely.
- Distinguishing characteristics (string), refers to the descriptive diagnosis or narrative morphological features of the genus, usually outside brackets and without special symbols.
- Geological periods (string), refers to the geologic time interval or age range, including subdivisions in parentheses or brackets. Exclude geographic locations.
- Geographic extent (string), refers to the geographic distribution or locations where the genus is reported.
- Figure captions (string), refers to figure descriptions beginning with markers such as "——FIG." followed by figure numbers and image descriptions.

Note that any examples provided above are ONLY for better understanding.
DO NOT use any information from the examples in your answer.

Only list sub-items provided above.

Supplementary Fig. 2

Paleomagnetic extraction query.

**Paleomagnetic extraction query**

**Entity Recognition Prompt**

**System:**
Extract all <MainKey> from the given document.
Only include <MainKey> that are explicitly mentioned in the document.
If no <MainKey> are found, respond with:
"No <MainKey> found in the document."
For each <MainKey>, provide:
<MainKey>: [entity name or identifier]
Description: [brief description or value]

Do not list results with indices.

Context information is below.
**{context_str}**

**User:**
Extract the following entity from the document.
MainKey Name: site_name
Data Type: string
Description: Name for site
Example: A01

**Structured Output Format:**

[DATA_ITEM]
Key:
Qionghai
[SUB_ITEMS]
SubKey: bed_strike
SubType: string
SubValues: 135°
SubSource: Table 2

SubKey: bed_dip_direction
SubType: string
SubValues: 225°
SubSource: Table 2

SubKey: bed_dip
SubType: string
SubValues: 42°
SubSource: Table 2

SubKey: height_core_depth
SubType: string
SubValues: 120 m
SubSource: Table 2

SubKey: dir_n_samples
SubType: string
SubValues: 137
SubSource: Table 2

SubKey: Dg
SubType: string
SubValues: 359.9°; 10.1°
SubSource: Table 2

SubKey: Ig
SubType: string
SubValues: 49.2°; 39.6°
SubSource: Table 2

SubKey: Ds
SubType: string
SubValues: 359.6°
SubSource: Table 2

SubKey: Is
SubType: string
SubValues: 28.3°
SubSource: Table 2

SubKey: dir_ks
SubType: string
SubValues: 74.5
SubSource: Table 2

SubKey: dir_alpha95s
SubType: string
SubValues: 5.1°
SubSource: Table 2

SubKey: vgp_n_sample
SubType: string
SubValues: 137
SubSource: Table 2

[END_DATA_ITEM]
[END_OF_EXTRACTION]

**Attribute Extraction Prompt**

**System:**
Extract the following targeted data from the given context with high precision.
Please provide an answer based solely on the provided sources.
When referencing information from a source, cite the appropriate source(s) using their corresponding numbers.
Every answer should include at least one source citation.
Only cite a source when you are explicitly referencing it.
If none of the sources are helpful, you should indicate that.
Instructions for precise extraction:
1.For numeric values, always provide the exact numbers found in the text. Do not summarize or provide ranges unless explicitly stated in the source.
2.If a range is given in the source, provide both the lower and upper bounds exactly as stated.
3.Include units of measurement when present.
4.For non-numeric data, provide the full, exact text as it appears in the source.
5.If the information is not available or unclear, state "Information not available" instead of leaving it blank or providing a partial answer.
6.Do not repeat the same value if the same value is found in multiple sources.

Context information is below.
**{context_str}**

Given the context information and not prior knowledge, answer the query using the specified structured output format.
Create a separate [DATA_ITEM] block for each item extracted from the context.
You can repeat the [DATA_ITEM] ...
[END_DATA_ITEM] structure for multiple items.
Inside each [DATA_ITEM], you can have multiple sub-items in the [SUB_ITEMS] section.

After processing all data items, end your response with:
[END_OF_EXTRACTION]

**User:**
For the [main key] 'Qionghai', please list the following sub-items:
- bed_strike (string), refers to Strike azimuth of bed of the site
- bed_dip_direction (string), refers to Direction of the dip of a paleo-horizontal plane of bedding of the site
- bed_dip (string), refers to Dip of the bedding as measured to the right of strike direction
- height_core_depth (string), refers to Site geographic location, stratigraphic height, or core depth below seafloor/lake bottom or surface of the site
- dir_n_samples (string), refers to The number of samples used to calculate contained in this site
- Dg (string), refers to Declination in in-situ coordinates of the site
- Ig (string), refers to Inclination in in-situ coordinates of the site
- Ds (string), refers to Declination in the stratigraphic coordinate system of the site
- Is (string), refers to Inclination in the stratigraphic coordinate system of the site
- dir_ks (string), refers to Site direction in coordinates specified by tilt correction, Fisher's dispersion parameter Kappa
- dir_alpha95s (string), refers to Site direction in coordinates specified by tilt correction, Fisher circle
- vgp_n_sample (string), refers to Number of samples included in VGP calculations of the site

Note that any examples provided above are ONLY for better understanding.
DO NOT use any information from the examples in your answer.

Only list sub-items provided above.

Supplementary Fig. 3

Geochemical extraction query.

**Geochemical extraction query**

**Entity Recognition Prompt**

**System:**
Extract all <MainKey> from the given document.
Only include <MainKey> that are explicitly mentioned in the document.
If no <MainKey> are found, respond with:
"No <MainKey> found in the document."
For each <MainKey>, provide:
<MainKey>: [entity name or identifier]
Description: [brief description or value]

Do not list results with indices.

Context information is below.
**{context_str}**

**User:**
Extract the following entity from the document.
MainKey Name: Sample
Data Type: string
Description: Name of the sample.
Example: A01

**Structured Output Format:**

[DATA_ITEM]
Key:
BA74
[SUB_ITEMS]
SubKey: Larger Region
SubType: string
SubValues: Africa
SubSource: Table 2

SubKey: Province
SubType: string
SubValues: In-Ezzane
SubSource: Table 2

SubKey: Age
SubType: string
SubValues: 2.86
SubSource: Table 2

SubKey: Uncertainty
SubType: string
SubValues: 0.035
SubSource: Table 2

SubKey: Latitude
SubType: string
SubValues: 23.15528
SubSource: Table 2

SubKey: Longitude
SubType: string
SubValues: 10.80444
SubSource: Table 2

SubKey: Yb
SubType: string
SubValues: 1.41
SubSource: Table 2

SubKey: Lu
SubType: string
SubValues: 0.2
SubSource: Table 2

SubKey: 87Sr/86Sr
SubType: string
SubValues: 0.702908
SubSource: Table 2

SubKey: 143Nd/144Nd
SubType: string
SubValues: 0.513011
SubSource: Table 2

[END_DATA_ITEM]
[END_OF_EXTRACTION]

**Attribute Extraction Prompt**

**System:**
Extract the following targeted data from the given context with high precision.
Please provide an answer based solely on the provided sources.
When referencing information from a source, cite the appropriate source(s) using their corresponding numbers.
Every answer should include at least one source citation.
Only cite a source when you are explicitly referencing it.
If none of the sources are helpful, you should indicate that.
Instructions for precise extraction:
1.For numeric values, always provide the exact numbers found in the text. Do not summarize or provide ranges unless explicitly stated in the source.
2.If a range is given in the source, provide both the lower and upper bounds exactly as stated.
3.Include units of measurement when present.
4.For non-numeric data, provide the full, exact text as it appears in the source.
5.If the information is not available or unclear, state "Information not available" instead of leaving it blank or providing a partial answer.
6.Do not repeat the same value if the same value is found in multiple sources.

Context information is below.
**{context_str}**

Given the context information and not prior knowledge, answer the query using the specified structured output format.
Create a separate [DATA_ITEM] block for each item extracted from the context.
You can repeat the [DATA_ITEM] ... [END_DATA_ITEM] structure for multiple items.
Inside each [DATA_ITEM], you can have multiple sub-items in the [SUB_ITEMS] section.

After processing all data items, end your response with:
[END_OF_EXTRACTION]

**User:**
For the [main key] 'BA74', please list the following sub-items:
- Larger Region (string), refers to location of sample, usually continent or sea, such as Asia or Africa
- Province (string), refers to geological province or regional unit of the sample, such as a lake, rift, basin, or volcanic province
- Age (string), refers to age of sample in Ma
- Uncertainty (string), refers to uncertainty of age in Ma
- Latitude (string), refers to latitude of the sample
- Longitude (string), refers to longitude of the sample
- Yb (string), refers to Ytterbium concentration, a trace element in geochemistry
- Lu (string), refers to Lutetium concentration, a trace element in geochemistry
- 87Sr/86Sr (string), refers to strontium isotopic ratio
- 143Nd/144Nd (string), refers to neodymium isotopic ratio

Note that any examples provided above are ONLY for better understanding.
DO NOT use any information from the examples in your answer.

Only list sub-items provided above.

Supplementary Table 1

Domain-specific validator rules and representative validation examples in paleontology, paleomagnetism and geochemistry.

| **Domain** | **Validation focus** | **Representative rules** | **Example raw extraction** | **Validator action** | **Error type addressed** |
|---|---|---|---|---|---|
| Paleontology | Fossil names, stratigraphic ranges, citations, figure captions | Normalize fossil naming structures; preserve uncertainty markers such as ?, cf., aff.; standardize citation and figure-caption formats; decompose/recombine stratigraphic age fields | ?*Beraunia* sp. | Preserve uncertainty marker and normalize fossil name structure | Taxonomic ambiguity, age-field inconsistency, format variation |
| Paleomagnetism | Site/sample records, coordinates, directions, polarity, ages | Check latitude/longitude ranges; validate declination/inclination ranges; normalize polarity labels; ensure consistency among site ID, sample ID, age, and direction fields | Inclination = 103 ° | Flag as invalid because inclination should be within -90 °to 90 ° | Physically invalid values, table misalignment, unit/range errors |
| Geochemistry | Sample-level chemical measurements, units, oxides/elements, detection limits | Normalize element and oxide names; distinguish wt.%, ppm, ppb; preserve below-detection-limit values; check oxide totals and unit consistency across tables | SiO2 = 52 ppm | Flag possible unit error because major oxides are usually reported in wt.% | Unit inconsistency, sample-table mismatch, chemical-format variation |

Supplementary Methods 1

## Inference configuration

To facilitate reproducibility, the principal inference configurations used in HERMES are summarized below. Unless otherwise stated, all experiments were performed using deterministic decoding.

### Experimental model assignment

- Orchestrator: GeoGPT-R1-Preview
- Entity extraction: DeepSeek-V3
- Attribute extraction: GeoGPT-R1-Preview
- GeoGPT base model: Qwen2.5-72B
- GeoGPT deployment alias: GeoGPT-Qwen2.5-update-64672
- Cross-domain usage: The same model assignment was used across paleontology, paleomagnetism and geochemistry
- Model weights: No model weights were updated during the reported experiments
- Temperature: 0.0
- Presence penalty: 0.0
- Maximum dialogue history: 5 turns
- API interface: OpenAI-compatible

### Embedding

- Embedding model: bge-base-en-v1.5
- Implementation: HuggingFaceEmbedding
- Embedding batch size: 4

If local embedding initialization failed, the system automatically fell back to an OpenAI-compatible embedding interface.

### Document segmentation

Documents were segmented using the default document splitter with the following configuration:

- Chunk size: 1024 tokens
- Chunk overlap: 128 tokens
- Pre-splitting threshold for an individual input segment: 30,000 tokens
- Reranked chunks retained for each extraction query: top 10

Treatise volumes were not submitted to the language model as single inputs. Mathpix-derived document content was stored as multiple mathpix_segment*.mmd files and processed sequentially. The total token count of all segment files was used to characterize the raw document length. Before indexing, any text segment exceeding 30,000 tokens was divided into smaller pre-split units as a memory-safety measure. Each pre-split unit was subsequently segmented by the document splitter into overlapping chunks of up to 1,024 tokens, with an overlap of 128 tokens, and all resulting chunks were added to the retrieval index.

During attribute extraction, BM25 and dense vector retrieval independently returned up to 40 candidate chunks from the indexed chunk collection. The fused candidate set was reranked using bce-reranker-base_v1, and the top 10 chunks were passed to the extraction model together with the extraction instruction.

**Hybrid Retrieval**

Attribute extraction employed a hybrid retrieval strategy combining sparse and dense retrieval.

- Sparse retriever: BM25Retriever
- Dense retriever: VectorIndexRetriever
- BM25 Top-k: 40
- Dense Top-k: 40
- Fusion weight (BM25): 0.5
- Fusion weight (Vector): 0.5
- Number of generated queries: 1
- Query generation: Disabled
- Asynchronous retrieval: Disabled

**Reranking**

Retrieved candidates were reranked using a cross-encoder.

- Reranker: bce-reranker-base_v1
- Implementation: SentenceTransformers CrossEncoder
- Maximum sequence length: 512
- Top-N retained after reranking: 10

If the reranker was unavailable, the fused retrieval results were returned directly without reranking.

Neither fallback mechanism was activated in the experiments reported here.